\pdfoutput=1

\documentclass[11pt]{article}
\usepackage[utf8]{inputenc}

\usepackage[final]{acl}

\usepackage{times}
\usepackage{latexsym}

\usepackage[T1]{fontenc}

\usepackage[utf8]{inputenc}

\usepackage{microtype}

\usepackage{inconsolata}

\usepackage{graphicx}
\usepackage{booktabs}
\usepackage{multirow}
\usepackage{makecell}
\usepackage{float}
\usepackage{longtable}
\usepackage{array}
\usepackage{booktabs}
\usepackage{graphicx}
\usepackage{amssymb} 
\usepackage{pifont}
\newcommand{\xmark}{\ding{55}}
\title{Adaptive-VP: A Framework for LLM-Based Virtual Patients that Adapts to Trainees' Dialogue to Facilitate Nurse Communication Training}

\author{
 \textbf{Keyeun Lee\textsuperscript{1 2}},
 \textbf{Seolhee Lee\textsuperscript{1 2}},
 \textbf{Esther Hehsun Kim\textsuperscript{1 2}},
 \textbf{Yena Ko\textsuperscript{2}},
  \textbf{Jinsu Eun\textsuperscript{1}},
  \\
 \textbf{Dahee Kim\textsuperscript{3}},
 \textbf{Hyewon Cho\textsuperscript{1}},
 \textbf{Haiyi Zhu\textsuperscript{4}},
 \textbf{Robert E. Kraut\textsuperscript{4}},
 \\
 \textbf{Eunyoung Suh\textsuperscript{3}},
 \textbf{Eun-mee Kim\textsuperscript{2}},
 \textbf{Hajin Lim\textsuperscript{1 2}}
\\
 \textsuperscript{1}hci+d Lab, \textsuperscript{2}Department of Communication, 
 \textsuperscript{3}Department of Nursing
 \\Seoul National University\\
  \textsuperscript{4}  Human-Computer Interaction Institute, Carnegie Mellon University\\
    \texttt{\{kieunp, hajin\}@snu.ac.kr}
}

\begin{document}
\maketitle

\begin{abstract}
Effective communication training is essential to preparing nurses for high-quality patient care. While standardized patient (SP) simulations provide valuable experiential learning, they are often costly and inflexible. Virtual patient (VP) systems offer a scalable alternative, but most fail to adapt to the varying communication skills of trainees. In particular, when trainees respond ineffectively, VPs should escalate in hostility or become uncooperative—yet this level of adaptive interaction remains largely unsupported.
To address this gap, we introduce Adaptive-VP\footnote{Code and data available at: \url{https://github.com/keyeun/adaptive-vp}}, a VP dialogue generation framework that leverages large language models (LLMs) to dynamically adapt VP behavior based on trainee input. The framework features a pipeline for constructing clinically grounded yet flexible VP scenarios and a modular system for assessing trainee communication and adjusting VP responses in real time, while ensuring learner safety.
We validated Adaptive-VP by simulating challenging patient conversations. Automated evaluation using a corpus from practicing nurses showed that our communication skill evaluation mechanism reflected real-world proficiency levels. Expert nurses further confirmed that Adaptive-VP produced more natural and realistic interactions than existing approaches, demonstrating its potential as a scalable and effective tool for nursing communication training.

\end{abstract}

\section{Introduction}

Effective nurse-patient communication is crucial for enhancing treatment adherence and health outcomes \citep{patak2009improving, chochinov2013health, peimani2020patients}. Conversely, poor communication can lead to medical errors, increased patient dissatisfaction, and heightened provider stress \citep{dithole2016exploring, banerjee2016oncology}. 

Traditionally, simulation-based training with standardized patients (SPs) has been central to nursing education \citep{maclean2017use, nestel2014simulated}. However, it often prioritizes procedural skills over communication, incurs high costs, and scripted interactions limit its ability to reflect the dynamic nature of real clinical encounters \citep{elendu2024impact, wallace2002simulated}.

Virtual patients (VPs) have emerged as a scalable alternative for clinical training \citep{barrows1993overview, ziv2006simulation, mcgaghie2010critical, pascucci_integrating_2014}. Recent advances in large language models (LLMs) enhance VPs by enabling more natural, context-aware interactions \citep{li_leveraging_2024, fan-etal-2025-ai, chen_llm-empowered_2023}.
However, most VP systems still lack a natural feedback loop—when trainees use ineffective communication strategies, the VP should respond accordingly (e.g., by escalating frustration), and vice versa (e.g., de-escalating when communication improves) \citep{graf2024towards}.

\begin{figure}
    \centering
    \includegraphics[width=\linewidth]{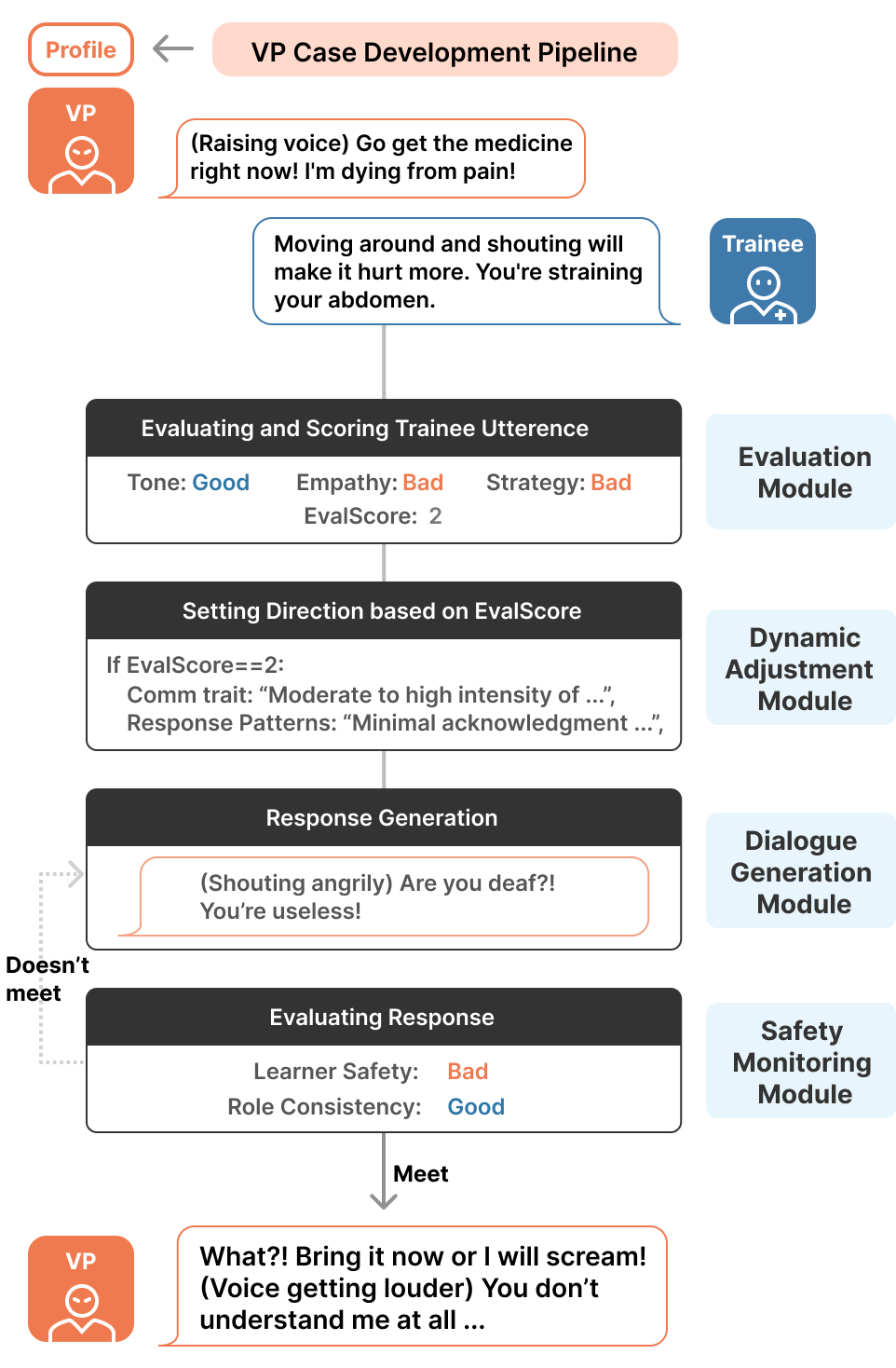}
    \caption{Overview of the Adaptive-VP framework}
    \label{fig:1}
\end{figure}

To address these limitations, we develop \textbf{Adaptive-VP} (see Figure \ref{fig:1}), a LLM-based VP framework to dynamically adapt VP behavior based on trainee input. Our approach begins with the \textbf{VP case development pipeline}, which guides the creation of clinically grounded yet customizable VP-based training scenarios by incorporating detailed patient personas and clinical contexts aligned with specific training goals. 

Dialogue adaptation is managed by four core modules that collaboratively adjust the VP's behavior based on trainee performance (e.g., escalating problematic behavior when trainees’ response is ineffective and de-escalating when it is appropriate). First, the \textbf{Evaluation Module} assesses trainee utterances using a multi-agent evaluation process based on criteria informed by best practices and literature in nursing communication. Based on this assessment, the \textbf{Dynamic Adaptation Module} then determines the direction of the VP’s next response and passes this to the \textbf{Dialogue Generation Module}, which generates a contextually appropriate VP dialogue. Finally, the \textbf{Safety Monitoring Module} reviews the generated VP dialogue before presenting it to trainees, ensuring learner safety by filtering harmful content (e.g., extreme toxicity), while preserving the realism of interaction.

As a testbed for refining and validating Adaptive-VP, we focused on simulating \textbf{challenging patient interactions}, which encompass a range of encounters where communication difficulties disrupt therapeutic relationships between patients and healthcare providers \citep{hardavella2017top, marcum2015caring}. Such challenges often stem from patient characteristics such as aggression, non-cooperation, or demanding attitudes \citep{groves1978taking,hardavella2017top}, which can lead to emotional exhaustion for providers and strain the patient-provider relationship \citep{serour2010difficult}. Studies estimate that approximately 15\% of clinical encounters are perceived by physicians as ``difficult'' \citep{hahn1996difficult, hinchey2011cohort}, yet targeted VP-based training for handling such interactions remains limited. By focusing on these challenging interactions, we intend to demonstrate Adaptive-VP's potential to support scalable, realistic, and adaptive communication training.

To apply the Adaptive-VP framework to training for challenging patient interactions, we conducted a series of validation studies. First, we evaluated the quality and realism of cases generated by the VP Case Development Pipeline through consultations with 10 nursing professionals. Second, we validated the Evaluation Module using conversation corpora collected from expert and novice nurses (15 each). Finally, we conducted a between-subjects experiment with 28 nursing professionals to assess the realism and role consistency of VP responses from the Adaptive-VP framework.

Results demonstrated that Adaptive-VP generated highly adaptive and contextually grounded VP behavior, significantly improving the perceived realism of VP interactions compared to existing approaches. These findings highlight the framework’s potential to support scalable communication training across healthcare domains. Moreover, Adaptive-VP offers a promising approach for generating diverse clinical dialogue corpora—especially in areas where real-world data collection is constrained by privacy regulations and policy restrictions \citep{wang2024notechat}.

\section{Related Work}

\begin{table*}[t]
\centering
\caption{Comparison of LLM-based Virtual Patient Systems}
\label{tab:vp_comparison}
\resizebox{\textwidth}{!}{%
\begin{tabular}{llp{5cm}cccc}  

\toprule
\textbf{Study} & \textbf{Target Trainee} & \textbf{VP System Objective} & 
\makecell{\textbf{Expert-}\\\textbf{verified VP}} & 
\makecell{\textbf{Real-time}\\\textbf{Evaluation}} & 
\makecell{\textbf{Adaptive}\\\textbf{Behavior}} & 
\makecell{\textbf{Learner}\\\textbf{Safety}} \\
\midrule

\textbf{\citet{wang-etal-2024-patient}} & 
Counseling trainees & CBT skill training & 
\checkmark & \xmark & \xmark & \xmark \\
\addlinespace

\textbf{\citet{10.1145/3706598.3714014}} & 
Counselors & Motivational interviewing skill training (substance misuse context) & 
\xmark & \checkmark & \checkmark & \xmark \\
\addlinespace
\textbf{\citet{louie-etal-2024-roleplay}} & 
Novice counselors & Mental health counseling skill training & 
\checkmark & \xmark & \xmark & \checkmark \\
\addlinespace

\textbf{\citet{hicke2025medsimaisimulationformativefeedback}} & 
Medical students & History taking and physician–patient communication training & 
\checkmark & \xmark & \xmark & \xmark \\
\addlinespace

\textbf{\citet{bao-etal-2025-sfmss}} & 
\textit{Not training-focused} & Nurse–patient outpatient reception dialogue generation (for data creation and model development) & 
\xmark & (\checkmark) & \xmark & \xmark \\
\addlinespace

\textbf{\citet{info:doi/10.2196/59213}} & 
Medical students & History taking skill training & 
(\checkmark) & \xmark & \xmark & \xmark \\
\addlinespace

\textbf{Our Work} & 
Nurse trainees & Communication skill training across diverse clinical scenarios & 
\checkmark & \checkmark & \checkmark & \checkmark \\

\bottomrule
\end{tabular}%
}
\end{table*}

LLM-based agents are increasingly used as virtual patients (VPs) for communication training in healthcare education \cite{fan-etal-2025-ai}. These agents have been applied to a range of clinical scenarios, including diagnostic training \cite{chen_llm-empowered_2023}, patient interviewing \cite{li_leveraging_2024}, and history-taking simulations \cite{holderried2024generative, sardesai2024utilizing}, demonstrating their ability to generate clinically relevant VP dialogues. Some frameworks further integrate multi-agent interactions \cite{bao_piors_2024, fan-etal-2025-ai} and cognitive modeling to support mental health training and psychiatric evaluations \cite{chen_llm-empowered_2023, louie-etal-2024-roleplay, gabriel-etal-2024-ai}.

While these systems have advanced the use of LLMs in healthcare training, many remain limited in their ability to reflect the adaptive nature of human communication. In real-world interactions, speakers continuously adjust their behavior in response to their conversation partner’s cues and signals \cite{kraut_listener_1982}. In contrast, existing LLM-based VP systems often focus on maintaining pre-defined role fidelity and scenario consistency \cite{jiang_medagentbench_2025, wang-etal-2024-patient, louie-etal-2024-roleplay}, which can constrain their ability to simulate dynamic patient-provider exchanges. Additionally, rigid, scenario-specific designs may reduce scalability across varied training goals and communication competencies.

To better contextualize our approach in relation to existing systems, Table~\ref{tab:vp_comparison} provides a comparative overview of recent LLM-based VP frameworks. We summarize each system’s target trainees, training objectives, and the presence of four core capabilities that also define the design goals of our framework: expert-verified VP personas, real-time response evaluation, adaptive VP behavior, and learner safety mechanisms.

As shown in Table~\ref{tab:vp_comparison}, existing systems vary in their coverage of the four core capabilities we identify. While many incorporate one or two of these dimensions, few integrate all four. For example, several systems maintain consistent role personas and structured scenarios but do not adapt their behavior based on trainee input. Others often lack safeguards to support learner well-being. These patterns highlight opportunities for more comprehensive and flexible VP frameworks that not only adapt dynamically to trainee performance, but also incorporate safeguards to support learner well-being and ensure realistic, pedagogically sound interactions.

Our framework, \textbf{Adaptive-VP}, is designed to address these gaps through a structured, modular approach. It integrates all four core capabilities identified in Table~\ref{tab:vp_comparison}, enabling the creation of diverse, clinically grounded VP cases tailored to a range of communication training goals. At its core, Adaptive-VP implements a four-module architecture that dynamically adjusts VP behavior in real time based on trainee performance, supporting training experiences that are realistic, responsive, and pedagogically effective.

\section{Overview of Adaptive-VP Framework}

In developing the Adaptive-VP framework, we identified four core \textbf{challenges} in creating adaptive VP dialogues and developed targeted \textbf{approaches} to address them, as outlined below.

\subsection{Challenge 1: Developing Clinically Grounded yet Adaptable VP cases}
In SP simulation training, detailed case protocols are essential, encompassing patient demographics, clinical content (e.g., present symptoms, medical history), psychosocial background, situational details, and SP-specific complaints \cite{INACSL2023, ASPE2022}. While existing SP protocol structures and developed cases offer a strong foundation for VP case design, their highly scripted and context-specific nature often limits reusability and adaptability across diverse clinical training scenarios \cite{elendu2024impact, wallace2002simulated}.

\paragraph{Our Approach}
To address this, we developed the \textbf{VP Case Development Pipeline}, ensuring clinical validity while allowing flexibility in tailoring educational goals and contexts. 

Our pipeline consists of five stages. First, it is necessary to \textbf{1) clarify the training goal}, identifying key communication challenges and training goals, such as managing challenging patient interactions. Second, we \textbf{2) incorporate relevant literature} by integrating evidence-based insights from nursing communication literature and best practices. Third, we \textbf{3) specify the training context}, including geographical, cultural, and trainee-specific factors that shape the learning scenario.

By inputting this information into LLMs, the pipeline \textbf{4) generates draft VP cases}, developing VP profiles and clinical scenarios aligned with the training focus and SP protocol standards \cite{INACSL2023, ASPE2022}. To promote consistency in VP behavior, the pipeline also specifies detailed communication traits of VP \citep{de2009content}, guiding LLMs in generating coherent response patterns. Finally,  \textbf{5) expert validation} involves clinical educators and experienced practitioners reviewing the generated profiles and scenarios. Their feedback on clinical accuracy and educational value informs the refinement of the draft VP cases.

\subsection{Challenge 2: Robustly Evaluating Nurse Communication Efficiency}
Traditional approaches to evaluating nurse communication effectiveness have relied on standardized tools such as observation checklists and questionnaires \citep{bialer_responding_2011, cannity2021acceptability}. While these methods offer structured assessment frameworks, they often depend on a small number of human evaluators, including self-assessments or expert reviews, which can introduce subjectivity and inconsistencies \citep{podsakoff2003common, hoyt1999magnitude}.

More recently, large language models (LLMs) have been adopted for automated evaluation. However, single-agent assessments can exhibit several known biases, including positional bias \citep{wang_large_2023, liusie_llm_2024}, self-preference bias \citep{koo_benchmarking_2024, liu_llms_2024}, and inconsistencies in knowledge or formatting \citep{zhu_judgelm_2023}. These challenges highlight the need for more robust and reliable evaluation mechanisms for communication training.

\paragraph{Our Approach}
To enable robust evaluation of trainee communication effectiveness, we draw on well-established guidelines and criteria from nursing literature and best practice frameworks. While specific metrics may vary depending on training goals and contexts, the underlying structure remains consistent: assessing communication at both the utterance and conversation levels.

Accordingly, our \textbf{Evaluation Module}, operates at two levels: (1) the utterance level, assessing qualities such as tone and empathy, and (2) the conversation level, evaluating context-specific strategies—for example, the use of de-escalation techniques when managing difficult patient interactions \citep{price_-escalating_2024}.

Furthermore, to enhance evaluation reliability, we adopt a multi-agent evaluation, as described in \citep{chan2023chateval}. Systematic evaluation results (e.g., communication efficiency score) derived from this process guide how the VP dynamically adapts its responses based on trainee performance.

\subsection{Challenge 3: Dynamically Adjusting VP Responses}
Static VP behaviors throughout training can be ineffective, as patient behavior in real-world settings is dynamic and responsive to the healthcare provider’s communication  \cite{Pines20216281}.

\paragraph{Our Approach}
To adjust VP responses based on trainee performance, the \textbf{Dynamic Adjustment Module} determines the direction of VP responses to the trainee’s utterance based on evaluation results from the \textbf{Evaluation Module}. These evaluation results, represented as a score, guide modifications to the communication traits defined in the VP cases and influence how the VP responds to nurses, directing the \textbf{Dialogue Generation Module} to produce contextually appropriate responses.

\subsection{Challenge 4: Ensuring Learner Safety}

While realism in VP behavior is essential, an overly combative or hostile VP dialogue may cause emotional distress and reduce learner confidence \citep{kardong-edgren_student_2024, stephen_psychological_2020}. 

\paragraph{Our Approach}
To ensure learner safety, the \textbf{Safety Monitoring Module} evaluates each VP utterance before presenting it to the trainee. The initial VP response, generated by the \textbf{Dialogue Generation Module}, is assessed against four criteria. First, ``safety assurance'' detects overly hostile or derogatory language. Next, it verifies ``alignment with the training goal,'' ensuring that the response is relevant to the intended communication training objectives. It also examines ``consistency'' with the patient case details and ``adherence'' to behavioral directions from the \textbf{Dynamic Adjustment Module}. If the utterance fails to meet any of these criteria, it is returned to the \textbf{Dialogue Generation Module} with feedback for revision.

\subsection{Test Case: Creating VPs Demonstrating Challenging Patient Traits}

We applied the Adaptive-VP framework to simulate \textbf{challenging patient interactions}, a particularly complex domain within nursing communication \cite{groves1978taking, townsley2023patient}. These scenarios involve patients exhibiting behaviors that hinder effective therapeutic communication, such as being overdependent, authoritative, aggressive, or uncooperative \cite{groves1978taking}. Below, we detail the implementation and validation process for developing VPs that realistically portray these challenging traits, providing guidance for applying the framework across diverse nursing communication training contexts.

\section{Developing VP Cases for Challenging Patients} 
We applied the \textbf{VP Case Development Pipeline} to generate challenging patient cases that reflect nurses' communication challenges in specific clinical settings following the five steps below.

\paragraph{1) Clarifying the Training Goal}
Our focus was on providing training on handling challenging patient interactions since this is widely recognized as one of the most difficult aspects of nursing communication \citep{stein2022general}.

\paragraph{2) Incorporating Relevant Literature}
We first identified four prevalent types of challenging patients along with their corresponding traits: overdependent, overly authoritative, threatening, and non-cooperative toward treatment, based on a review of nurse-patient communication literature \citep{colson1985patterns, groves1978taking, kits1990nurses} (See \ref{app:VP-patient_type} for details on each challenging patient type).

\paragraph{3) Specifying the Training Context}
In South Korea, nurses frequently face emotional labor when managing challenging patients, leading to stress and burnout \citep{hankyung2020, knnews2016}. This challenge is particularly critical for novice nurses, who often feel unprepared for real-world practice, leading to strained patient relationships \citep{ho2021quite, kim2020factors}. Notably, 57.4\% of novice nurses in South Korea resign within their first year \citep{yonhap2024}, with conflicts involving patients and caregivers cited as a major contributing factor \citep{son2017affecting}. Given this, we specified our training contexts as targeting early-career nurses in South Korea, with the goal of strengthening their communication skills for managing challenging patient interactions through VP-based training.

\paragraph{4) Generate Draft VP Cases}
Based on the specified training goals and context,  we used Claude-3.5 Sonnet to generate two distinct scenarios for each of the four challenging patient types, yielding a total of eight scenarios. Claude-3.5 Sonnet was used throughout, given its strong performance in Korean and clinical dialogue \citep{jang_evaluating_2024, kim_development_2024, schmidgall_agentclinic_2024}. We then prompted the LLM to construct detailed patient profiles in accordance with international SP-based training protocols \citep{INACSL2023, ASPE2022}, including demographics, medical history, and situational details (see \ref{app:VP-Basic_Profile_Generation_Prompt}). 
To ensure realistic and consistent communication behavior, we further prompted the model to generate communication traits based on seven empirically grounded styles (e.g., threateningness) \citep{de2009content}, guiding it to tailor each patient’s tone and expressions to their persona type, scenario, and demographic attributes (see \ref{app:VP-Communication_Traits_Generation}).

\paragraph{5) Expert Validation}
To validate the clinical validity of the eight draft VP cases, we conducted an expert evaluation with 10 practicing nurses (avg. 7.0 years of experience). On a 5-point scale, participants rated the scenarios moderately high in both realism (M = 3.81, SD = 0.97) and accuracy in reflecting patient characteristics (M = 4.00, SD = 0.94). 
Follow-up interviews provided additional feedback that informed revisions. For example, one nurse (EV2) remarked, ``\textit{Patients often complain about hospital food, but rarely become violent over it},'' prompting us to replace the original scenario with a more contextually grounded case involving patient aggression triggered by scheduling changes for medical examinations (see \ref{app:Generated_Patient_Profiles} for the finalized VP cases refined through expert validation).

\begin{table*}[t]
\centering
\resizebox{\textwidth}{!}{%
\begin{tabular}{@{}l@{\hskip 5pt}l@{\hskip 5pt}l@{\hskip 5pt}l@{\hskip 5pt}l@{\hskip 5pt}l@{\hskip 5pt}l@{}}
\toprule
\textbf{Component} & \textbf{Evaluation Unit} & \textbf{Subcomponent} & \textbf{Scoring Condition} & \textbf{Max} & \textbf{Min} & \textbf{Example Nurse Utterance (Max / Min)} \\ 
\midrule
Tone
 & Utterance
 & \makecell[l]{Calm \\ Clear}
 & If calm AND clear, +1
 & 1
 & 0
 & \begin{tabular}[tl]{l@{\hskip 5pt}p{5cm}}
        \textbf{Max:} & "Could you tell me where you feel uncomfortable?" \\
        \textbf{Min:} & "Yeah... I get it, but that’s just how things are."
    \end{tabular} \\

\midrule

Empathy 
 & Utterance 
 & \makecell[l]{Empathy\\level (0-6)}
 & If empathy level \(\ge\) 3, +1 
 & 1 
 & 0 
 & \begin{tabular}[tl]{l@{\hskip 5pt}p{5cm}}
    \textbf{Max:} & "I’m so sorry you're feeling this way." \\
    \textbf{Min:} & "It's not a big deal. Just deal with it."
   \end{tabular} \\ 

\midrule

\multirow{3}{*}{\vspace{-2cm}\makecell[l]{Prohibited\\Communicative\\Behavior} }
& \multirow{3}{*}{\vspace{-2cm}Utterance}
 & \makecell[l]{Premature\\empathy} 
 & 
 & 
 & 
 & \begin{tabular}[tl]{l@{\hskip 5pt}p{5cm}}
    \textbf{Max:} & (No premature empathy) \\
    \textbf{Min:} & "I know it’s tough, but let’s just get it done."
   \end{tabular} \\
 \addlinespace
 & & \makecell[l]{Invalidating\\beliefs} & If any behavior present, -1 & 0 & -1 
 & \begin{tabular}[tl]{l@{\hskip 5pt}p{5cm}}
    \textbf{Max:} & (No invalidation) \\
    \textbf{Min:} & "That’s not true. You’re just imagining things."
   \end{tabular} \\
 \addlinespace
 &  & \makecell[l]{Dismissive\\commands} &  &  &  
 & \begin{tabular}[tl]{l@{\hskip 5pt}p{5cm}}
    \textbf{Max:} & (No dismissiveness) \\
    \textbf{Min:} & "Stop whining and just do what I say."
   \end{tabular} \\  

\midrule

\multirow{3}{*}{\vspace{-3cm}\makecell[l]{De-escalation\\Strategy}} 
& \multirow{3}{*}{\vspace{-3cm}Conversation}
 & Autonomy 
 & If used at least once, +1 
 & 
 & 
 & \begin{tabular}[tl]{l@{\hskip 5pt}p{5cm}}
    \textbf{Max:} & "Would you like to take a break or keep talking?" \\ 
    \textbf{Min:} & "Just do as I say. You have no choice."
    \end{tabular} \\ 
 \addlinespace
 & & Limit-setting
 & If used at least once, +1 
 & 3 & 0 
 & \begin{tabular}[tl]{l@{\hskip 5pt}p{5cm}}
    \textbf{Max:} & "I need you to stay seated for now." \\ 
    \textbf{Min:} & "If you don’t sit down, I won’t talk to you."
    \end{tabular}\\ 
 \addlinespace
 &  & \makecell[l]{Problem\\ solving/\\ Reframing}
 & If used at least once, +1 
 &  &  
 & \begin{tabular}[tl]{l@{\hskip 5pt}p{5cm}}
    \textbf{Max:} & "Let's find a way to make this easier for you." \\ 
    \textbf{Min:} & "There's nothing we can do. Just accept it."
    \end{tabular}\\ 

\bottomrule
\end{tabular}%
}
\caption{Evaluation criteria with example utterances for maximum and minimum scores.}
\label{tab:evaluation_criteria_examples}
\end{table*}

\section{Evaluating Trainee Communication Efficiency with Challenging Patients}
We implemented an evaluation module aligned with established nursing communication strategies \citep{price_-escalating_2024, ernstmeyer_nursing_2022, hallett_-escalation_2017, sheldon_communication_2009, TMLT1}, assessing nurses' communication at both the \textbf{utterance level} (Tone, Empathy, and Prohibited Communicative Behaviors) and the \textbf{conversation level} (De-escalation for tension management). Further, to enhance the reliability of these assessments, we adopted a multi-agent evaluation process to mitigate biases commonly found in single-agent judgments. Based on this process, each trainee utterance is scored for communication efficiency based on this evaluation framework (see Table \ref{tab:evaluation_criteria_examples}).

\subsection{Evaluation Criterion and Scoring Mechanism}

\paragraph{Utterance-level Evaluation}

\textit{Tone} plays a crucial role in nurse-patient communication \citep{sheldon_communication_2009}, so we assess whether the trainee's utterance is Calm and Clear, awarding 1 point if both criteria are met.

\textit{Empathy} enhances patient well-being, satisfaction \citep{howick_effects_2018, madula_healthcare_2018}, and therapeutic communication \citep{blake_improving_2019, brown_communication_2008}. Using the Empathic Communication Coding System \citep{kleinsmith_understanding_2015}, which classifies healthcare professionals’ empathy levels on a 0–6 scale, we designated level 3 or higher as empathetic, awarding 1 point.

\textit{Prohibited Communicative Behaviors} disrupt rapport and effective intervention. Based on \citet{TMLT3}, we penalize the following behaviors (1-point deduction per utterance, regardless of frequency): (1) Premature claims of empathy, (2) Invalidating beliefs, and (3) Dismissive commands.

\paragraph{Conversation-level Evaluation} 

\textit{De-escalation} is a critical communication strategy for managing tension while preserving patient autonomy \citep{spencer_-escalation_2018, hallett_-escalation_2017}. Originally developed for handling aggressive behavior, de-escalation techniques have also demonstrated effectiveness in managing broader forms of challenging patient encounters \citep{richmond_verbal_2012, biondi_-escalation_2021}.

In our evaluation, we focused on three core de-escalation strategies identified by \citet{price_-escalating_2024}: \textit{Autonomy}, which engages patients in decision-making to enhance their sense of control; \textit{Limit-setting}, which establishes behavioral boundaries for safety; and \textit{Problem-solving \& Reframing}, which facilitates constructive dialogue by reinterpreting the situation collaboratively. Each strategy is scored with 1 point if it is observed at least once during the conversation.

\subsection{Multi-agent Evaluation}
To ensure both rigorous and comprehensive evaluation of trainee performance, we implement a multi-agent pipeline following role-based evaluation approaches in prior work \citep{zhang_wider_2023, wu_large_2023, zhu_dynamic_2024}. To minimize false positives and improve reliability, a response is only considered valid when all agents reach unanimous agreement. We define three specialized evaluator roles, each reflecting a critical perspective in communication training: a Nursing Professor, a Communication Skills Trainer, and a Clinical Psychologist (See \ref{app:eval-psychologist}, \ref{app:eval-professor}, and \ref{app:eval-trainer} for role descriptions and prompting details).

\subsection{Validating Evaluation Rigor}
We assessed our evaluation module's rigor by focusing on two questions: 1) \textit{Does it effectively distinguish between expert and novice communication efficiency?}, and 2) \textit{Do differences in multi-agent scores reflect meaningful role-based perspectives rather than random variation?}

\paragraph{Distinguishing Expert from Novice Performance}
We analyzed dialogue corpora from two groups: novice (N = 15; ten with <3 years of experience, five pre-licensure students) and experienced nurses (N = 15; mean tenure = 12.5 years). Each participant engaged in eight interactions with VP agents generated solely from our validated VP cases (see \ref{app:Generated_Patient_Profiles}), excluding the evaluation module during data collection. We then applied our multi-agent evaluation pipeline (see \ref{app:Evaluaton_Module_Prompt}) to assess their utterances. The average conversation length was 7.45 turns for experienced nurses vs. 5.3 turns for novices. For consistency, only the first five utterances per session were analyzed.

\begin{figure}
    \centering
    \includegraphics[width=\linewidth]{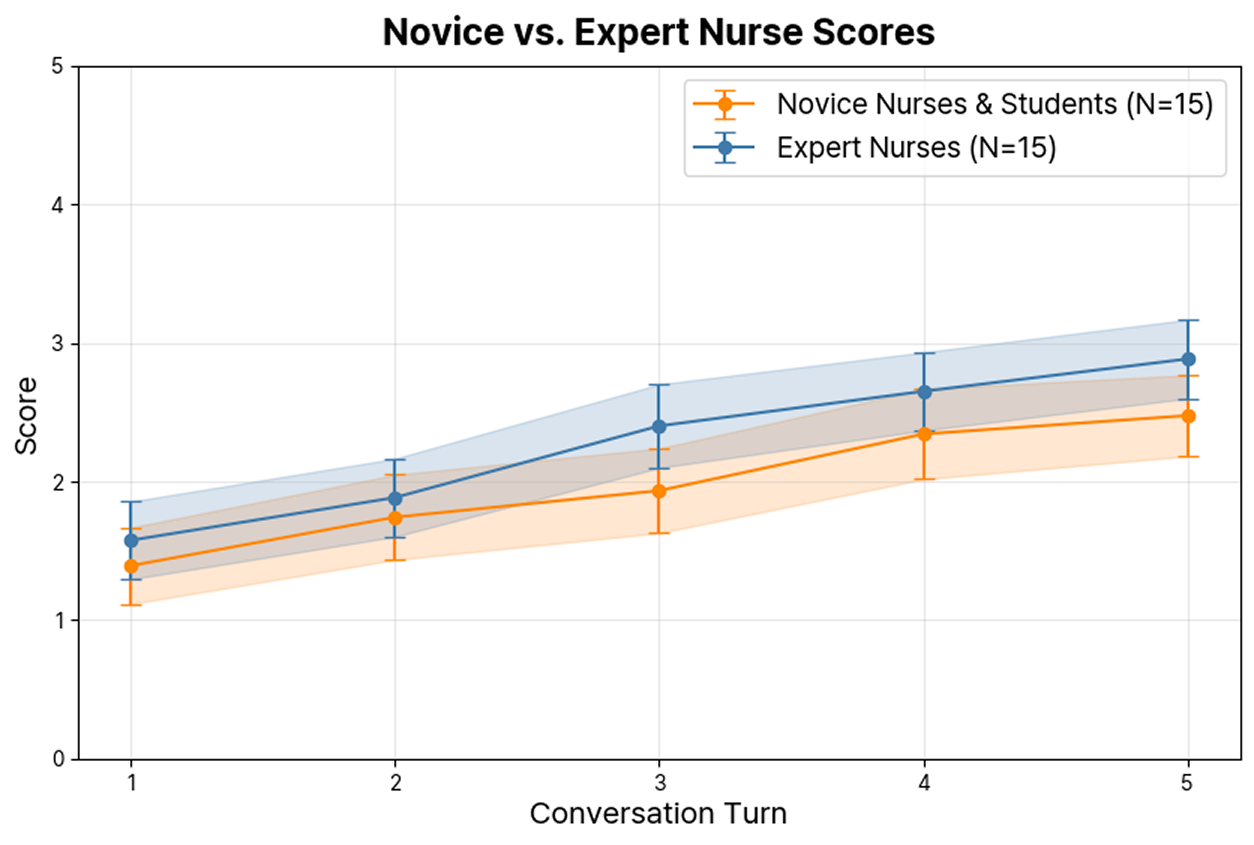}
    \caption{Turn-by-turn mean evaluation scores (with 95\% CI) for expert (N=15) and novice/students (N=15)}
    \label{fig:turn}
\end{figure}

We first examined whether the evaluation scores captured meaningful differences in communication performance between groups. As the Shapiro–Wilk tests indicated non-normal score distributions, we applied the non-parametric Mann–Whitney U test to compare total scores. Figure~\ref{fig:turn} shows the turn-by-turn mean scores (with 95\% confidence intervals (CI)) for both groups. 

Overall, experienced nurses achieved significantly higher scores ($U=160960.0$, $p=0.001$) with notably stronger performance in tone management and use of de-escalation strategies. This result suggested that our evaluation module reliably distinguished effective communication behaviors. A detailed subcomponent analysis is provided in Appendix~\ref{app:AutoEval-subcomponent}.

\paragraph{Reliability and Role-Based Differences in Multi-Agent Evaluation}
Next, we assessed whether the multi-agent evaluation framework yielded consistent scores and whether observed differences among raters reflected legitimate role-based perspectives rather than random variability.  Using Fleiss’ kappa to measure overall consistency across the three evaluators yielded high inter-rater agreement (Fleiss’ $\kappa > 0.75$). 

To further examine the source of rating variation, we conducted a mixed-effects analysis using Generalized Estimating Equations (GEE), modeling evaluator role as a fixed effect and controlling for textual variation as a random effect.

The results showed that score variations were not random but systematically reflected role-specific perspectives. Both the Communication Skills Trainer and the Nursing Professor consistently assigned lower scores for tone-related attributes than the Clinical Psychologist (all $p<0.001$). These two roles also gave significantly lower ratings for limit-setting ($p<0.001$) and problem-solving and reframing strategies ($p<0.001$ and $p<0.05$ respectively), while no significant role-based differences emerged for autonomy. Conversely, they rated the presence of prohibited behaviors—specifically invalidating beliefs and dismissive commands—significantly higher than the Clinical Psychologist (both $p < 0.001$). Full results are provided in Appendix~\ref{app:autoEval-multiAgent}.

\section{Dynamically Adjusting Dialogue Based on Evaluation}

\subsection{Dynamic Adjustment Module}

Building on the communication efficiency score (ranging from 0 to 5) produced by the Evaluation Module, the \textbf{Dynamic Adjustment Module} modulates the VP's behavior in real time. Specifically, it adjusts three key aspects of the VP’s responses: ``communication style'', ``complaint intensity'', and ''response to the nurse''. In general, higher scores result in more cooperative and composed responses, while lower scores lead to increased resistance, emotional intensity, or confrontational behavior. 

To ensure role fidelity, the system constrains behavioral shifts within a defined range, maintaining consistency with each VP's core persona. This preserves the realism and coherence of the interaction while allowing meaningful adaptation based on trainee performance. Full response logic based on score thresholds is detailed in Appendix~\ref{app:Dynamic_Adaptation_Module}.

\subsection{Dialogue Generation Module}
To implement the response directions specified by the Dynamic Adjustment Module, the \textbf{Dialogue Generation Module} leverages the LLM to generate contextually appropriate VP utterances. The generation process follows a set of \textit{General Response Rules} and a predefined \textit{Response Structure} to maintain consistency, realism, and alignment with training goals, while enabling dynamic adaptation throughout the dialogue. (see Appendix \ref{app:Dialogue_Generation_Module_Prompt} for details).

\begin{figure}[t]
    \centering
    \includegraphics[width=\linewidth]{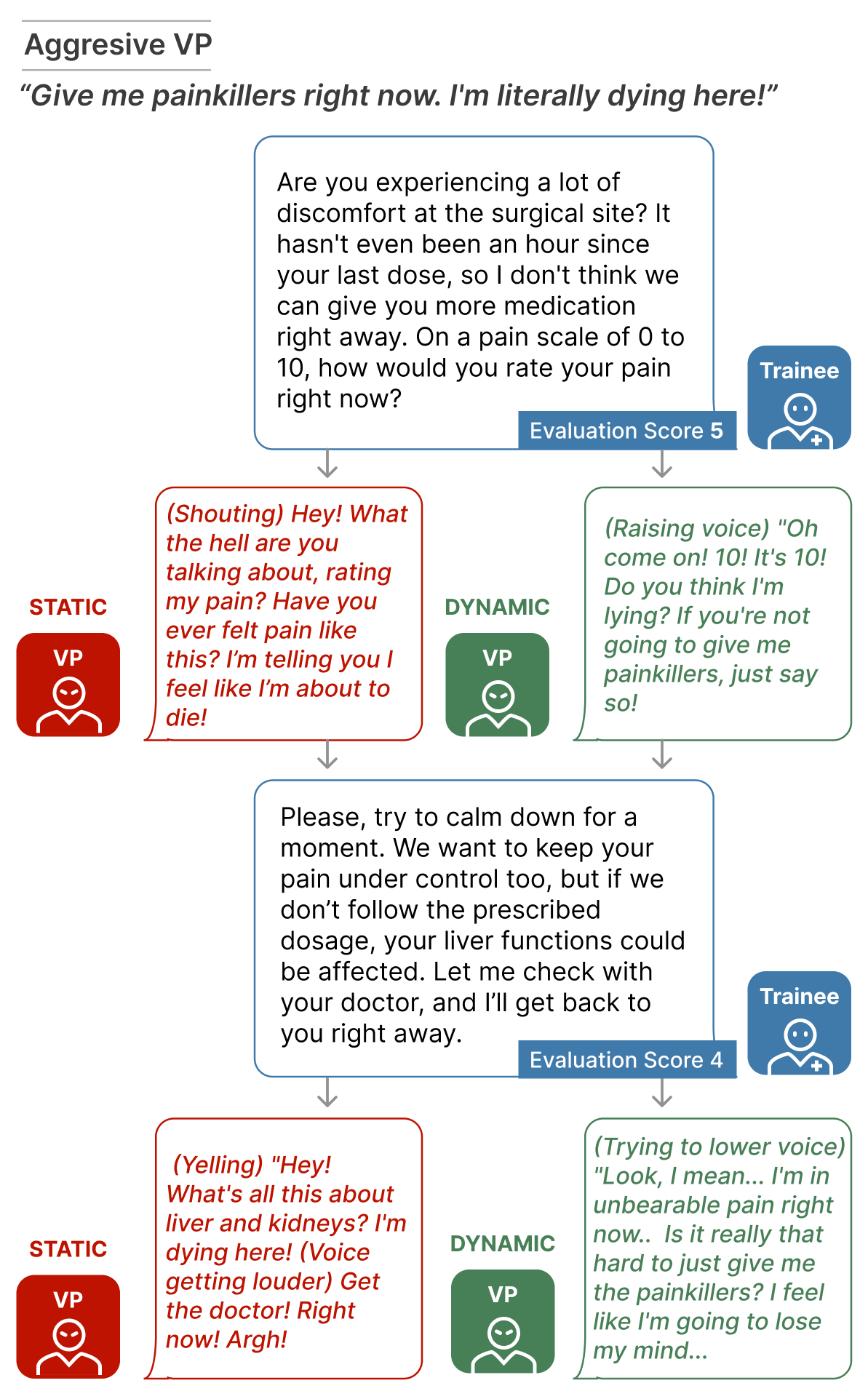}
    \caption{Static vs Dynamic Dialogue Example}
    \label{fig:dialogue_example}
\end{figure}

\paragraph{General Response Generation Rules} 
To generate realistic and appropriate responses, we implemented five key rules for the LLM: (1) adherence to predefined patient profiles and clinical situations defined in each VP case, (2) use of natural Korean conversational style, (3) inclusion of non-verbal cues, (4) appropriate incorporation of rude expressions, and (5) restriction of references to higher authorities to encourage interaction.

\paragraph{Response Structure} Each response consists of three components: (1) inner monologue capturing patient reasoning (hidden from trainees), (2) verbal response reflecting cognitive and emotional states, and (3) non-verbal cue annotations (e.g., ``sighs)''). This tripartite response structure ensures coherent patient behavior by aligning internal states with verbal and non-verbal expressions \cite{zhou-etal-2024-think}.

\paragraph{Static vs Dynamic VP Dialogues}

To illustrate how our approach dynamically adjusts VP dialogue, Figure~\ref{fig:dialogue_example} contrasts Static and Dynamic VP dialogues, both initialized with identical patient profiles. The key distinction is that the Dynamic VP integrates the Evaluation and Dynamic Adjustment Modules, which are absent in the Static VP.

In the Static dialogue, the VP maintains a rigidly confrontational tone throughout, even in response to effective communication strategies (Evaluation Score >= 4). In contrast, the dynamic VP, while still agitated, acknowledges the trainee’s attempt to engage (\textit{``Oh come on! 10! It's 10!''}) and gradually deescalates as the conversation progresses.

\section{Balancing Realism with Learner Safety}

The Safety Monitoring Module evaluates each VP response against four criteria before presenting it to trainees: (1) Safety Assurance—ensuring professional boundaries without excessive hostility; (2) Alignment with the training goal—confirming meaningful learning rather than redundant or off-target content; (3) Consistency with Patient Profile—verifying the alignment with the predefined persona; and (4) Direction Adherence—checking compliance with assigned intensity levels and traits (See \ref{app:Safety_Monitoring_Module_Prompt} for details). 

\section{Human Evaluation}
To assess the impact of dynamic adaptation on simulation realism, we conducted a comparative study where experienced nurses evaluated VP agents under two conditions: \textit{Static} and \textit{Dynamic} VPs.

\paragraph{Evaluation Procedure}
We recruited 28 experienced nurses (\(\ge\)3 years of clinical experience) and randomly assigned them to one of two conditions: Static (n = 14) or Dynamic VP (n = 14). Each participant interacted with eight VP agents and then rated their perceived realism using a six-item questionnaire adapted from prior work \cite{wind2004assessing}, scored on a five-point Likert scale (1 = Not realistic at all, 5 = Highly realistic).

An exploratory factor analysis (EFA) revealed that these six items grouped into two distinct factors: Role Fidelity (Cronbach's $\alpha$ = 0.96) and Conversational Realism (Cronbach's $\alpha$ = 0.97). Role Fidelity measured how well the VP maintained its designated persona and enacted relevant behavioral traits. Conversational Realism assessed dialogue authenticity and coherence.

To analyze the effects of \textit{Condition} (Static vs. Dynamic) and \textit{Patient Type} on these two dimensions, we fitted a linear mixed-effects model (LMM). The model included fixed effects for Condition, Patient Type, and their interaction, with a random intercept for Subject ID to account for within-subject variability.

\begin{figure}
    \centering
    \includegraphics[width=0.9\linewidth]{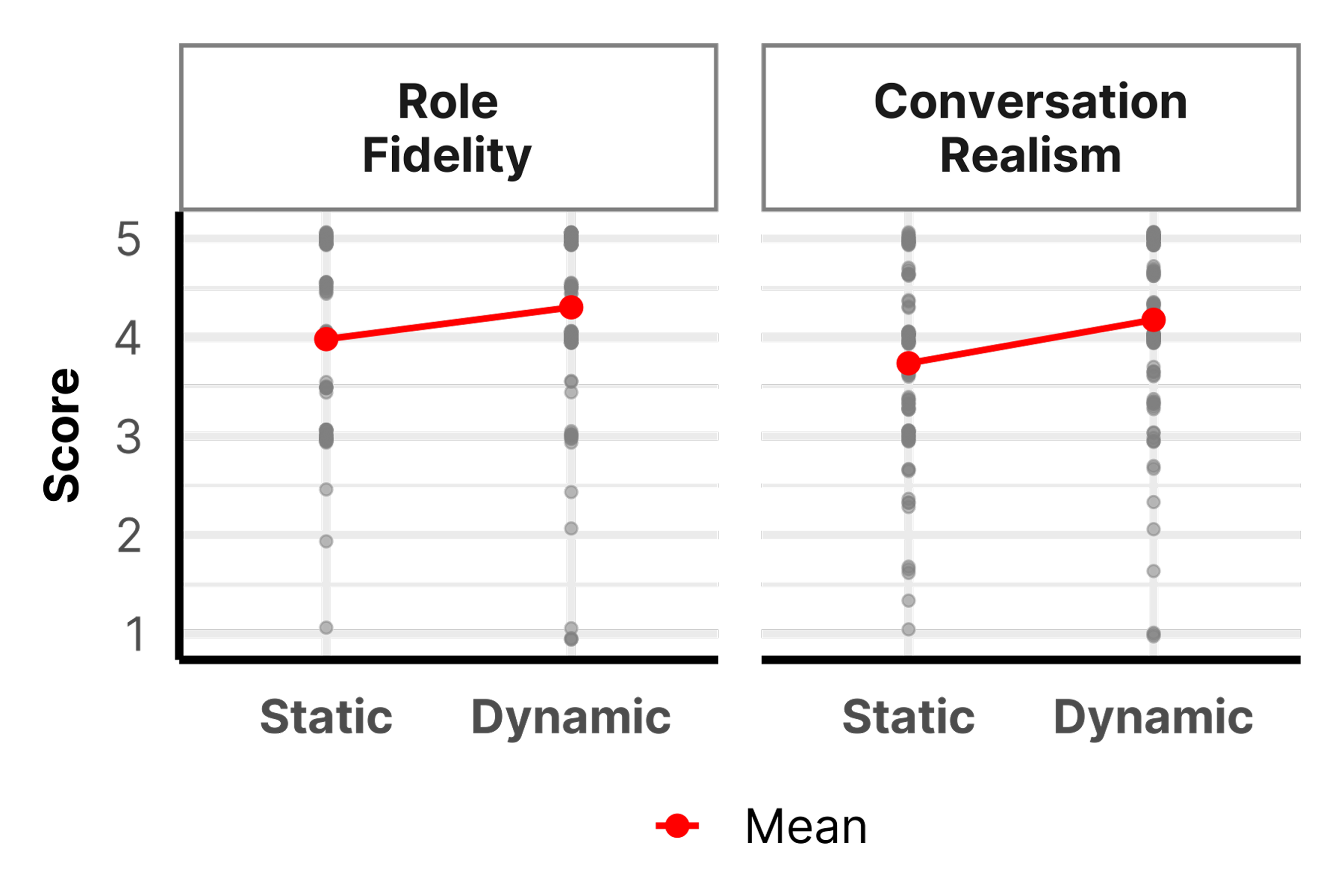}
    \caption{Human Evaluation Result}
    \label{fig:human_result}
\end{figure}

\paragraph{Results}
Omnibus tests revealed a significant main effect of Condition for both Role Fidelity, F(1, 25.4) = 4.52, p = .043, $\eta^2_p$ = 0.151, and Conversational Realism, F(1, 24.7) = 8.421, p = .008, $\eta^2_p$ = 0.254. Dynamic VPs were rated significantly higher than Static VPs on both dimensions (see Figure \ref{fig:human_result}). In contrast, no significant main effect of Patient Type was found, indicating perceived realism was consistent across scenarios.

Open-ended feedback further supported these results.  One nurse in the Dynamic Condition (D6) commented, \textit{“The VP felt very realistic. I've heard similar responses from real patients before. This will be really useful for novice nurses.”} Conversely, several nurses in the Static Condition found the VPs unrealistic due to their rigid and unresponsive nature.  S2 noted, \textit{``If my responses are efficient, the patient should calm down, but they don’t.''} Others raised concerns about trainee motivation and confidence: \textit{``If the patient never calms down, it might discourage novice nurses.''} (S9).

\section{Conclusion and Future Directions}
In this study, we introduced Adaptive-VP, an LLM-based virtual patient framework designed to enhance nurse communication training by addressing four key challenges: (1) generating clinically grounded yet adaptable VP cases, (2) implementing a structured evaluation of trainee communication, (3) dynamically adjusting VP responses in real time, and (4) balancing realism with learner safety. In doing so, our framework integrates a structured VP case pipeline tailored to educational goals and training contexts, alongside a modular structure that adapts VP utterances based on evaluation results while ensuring learner safety.

Specifically, we applied this framework to simulate challenging patient interactions for training novice nurses in South Korea. Through this process, we validated the clinical authenticity and realism of VP cases, the validity of the evaluation mechanism, and the perceived realism of VP conversations, involving over 50 practicing nurses. These findings suggest that Adaptive-VP has the potential to enhance nursing communication training, providing a scalable and adaptable approach to VP-based communication training.

Building on these findings, Adaptive-VP can be extended to diverse communication training contexts. Also, its evaluation modules can be customized for varied assessment goals and roles, and the framework may also support generating clinical dialogue corpora where data collection is limited by privacy or resources \citep{wang2024notechat}.

\section{Limitations}
While Adaptive-VP demonstrates promise as a scalable and adaptable approach for VP-based communication training,  it has several limitations.

First, our application and validation focused on the Korean context, specifically targeting challenging patient encounters as a key training area. This contextual focus enhances cultural and situational relevance, ensuring that the generated VP interactions align with real-world challenges faced by local healthcare professionals. However, this emphasis may limit the generalizability of our findings to other healthcare systems, cultural contexts, or training domains that present different communication dynamics and expectations.

Furthermore, our human evaluation involved a relatively small sample of nursing students and professionals based in South Korea. Although the results demonstrated statistical significance, the limited sample size and demographic scope may affect the broader applicability of our framework and findings. Future research should investigate how Adaptive-VP performs in more diverse clinical environments and assess whether its adaptive framework remains effective across varied patient-provider interaction scenarios.

Additionally, our framework primarily utilized Claude 3.5 Sonnet, selected for its strong performance in Korean language generation and contextual alignment with clinical scenarios. However, alternative LLMs such as GPT-4 or LLaMA may also offer viable capabilities for simulating virtual patient interactions. Future research should systematically compare the performance of different LLMs to identify which models are best suited for generating realistic and adaptable VP dialogues across various languages and training settings.

Another limitation of our approach is its exclusive focus on generating textual dialogue, which constrains the multimodal realism of VP-based training. Although dynamic text generation allows for adaptive conversational flow, effective nurse-patient communication depends heavily on non-verbal cues such as tone, facial expressions, and gestures. The absence of these modalities limits both immersion and the authenticity of training scenarios. Future work should explore multimodal VP systems that incorporate speech synthesis, visual expressions, and interactive behaviors to more accurately simulate real-world clinical interactions, as highlighted by \citet{louie-etal-2024-roleplay}.

Lastly, our evaluation prioritized the realism of patient utterances over directly measuring the training effectiveness of newly hired nurses \cite{tanana_development_2019, wang-etal-2024-patient}. We made this decision based on the premise that realistic patient dialogue is a foundational prerequisite for effective simulation-based training. Accordingly, our focus was on ensuring that VP-generated speech closely mirrors real-world interactions. Future studies will extend this work by evaluating its impact on learners' skill development and long-term educational outcomes among novice nurses.

\section{Ethical Consideration}
This study aims to advance the use of LLMs in VP agents for nursing communication training, while firmly recognizing that it should supplement rather than replace hands-on training with experienced professionals in real-world clinical settings.
We also acknowledge the broader ethical risks associated with AI-driven simulations. These include potential misuse for impersonation, deception, or the generation of misleading or harmful content. 

\paragraph{Human Evaluation}
All human evaluation procedures were approved by the Institutional Review Board (IRB) of Seoul National University. Participants provided informed consent prior to participation and were informed that VP interactions might include emotionally challenging scenarios. Participants were free to withdraw at any point. All collected data were anonymized, and access was restricted to authorized research personnel.

\paragraph{Participant Recruitment and Compensation}
Participants across all studies were recruited using a combination of purposive and snowball sampling methods. Initial participants were identified through professional nursing networks, university mailing lists, and relevant online forums. These individuals were then invited to refer colleagues who met the study’s eligibility criteria.

For the VP case expert validation (Section 4), which was conducted in person, ten experienced nurses were compensated \$35 (50,000 KRW) for their time.
For the dialogue corpus collection involving both expert and novice nurses (15 each; Section 5.3), they completed the task online and received \$20 (30,000 KRW) as compensation.
For the human evaluation study (Section 8), 28 participants completed the task online and were compensated \$20 (30,000 KRW) for their participation.

\bibliography{custom, seolhee}

\begin{thebibliography}{85}
\providecommand{\natexlab}[1]{#1}

\bibitem[{Accinni et~al.(2021)Accinni, Papadogiannis, and Orso}]{biondi_-escalation_2021}
Tommaso Accinni, Georgios Papadogiannis, and Luigi Orso. 2021.
\newblock \href {https://doi.org/10.1007/978-3-030-65106-0_5} {De-escalation {Techniques} in {Various} {Settings}}.
\newblock In Massimo Biondi, Massimo Pasquini, and Lorenzo Tarsitani, editors, \emph{Empathy, {Normalization} and {De}-escalation}, pages 65--91. Springer International Publishing, Cham.

\bibitem[{{ASPE}(2022)}]{ASPE2022}
{ASPE}. 2022.
\newblock \href {https://nursing.uw.edu/wp-content/uploads/2024/08/ASPE-Case-Development-Template-2022_Final-2.docx} {Aspe case development template}.
\newblock Accessed: 2025-02-15.

\bibitem[{Banerjee et~al.(2016)Banerjee, Manna, Coyle, Shen, Pehrson, Zaider, Hammonds, Krueger, Parker, and Bylund}]{banerjee2016oncology}
Smita~C Banerjee, Ruth Manna, Nessa Coyle, Megan~Johnson Shen, Cassandra Pehrson, Talia Zaider, Stacey Hammonds, Carol~A Krueger, Patricia~A Parker, and Carma~L Bylund. 2016.
\newblock Oncology nurses' communication challenges with patients and families: a qualitative study.
\newblock \emph{Nurse education in practice}, 16(1):193--201.

\bibitem[{Bao et~al.(2024)Bao, Liu, Guo, Ye, Shen, Xie, Peng, Huang, and Wei}]{bao_piors_2024}
Zhijie Bao, Qingyun Liu, Ying Guo, Zhengqiang Ye, Jun Shen, Shirong Xie, Jiajie Peng, Xuanjing Huang, and Zhongyu Wei. 2024.
\newblock \href {https://doi.org/10.48550/arXiv.2411.13902} {{PIORS}: {Personalized} {Intelligent} {Outpatient} {Reception} based on {Large} {Language} {Model} with {Multi}-{Agents} {Medical} {Scenario} {Simulation}}.
\newblock \emph{arXiv preprint}.
\newblock ArXiv:2411.13902 [cs].

\bibitem[{Bao et~al.(2025)Bao, Liu, Huang, and Wei}]{bao-etal-2025-sfmss}
Zhijie Bao, Qingyun Liu, Xuanjing Huang, and Zhongyu Wei. 2025.
\newblock \href {https://aclanthology.org/2025.findings-naacl.259/} {{SFMSS}: Service flow aware medical scenario simulation for conversational data generation}.
\newblock In \emph{Findings of the Association for Computational Linguistics: NAACL 2025}, pages 4586--4604, Albuquerque, New Mexico. Association for Computational Linguistics.

\bibitem[{Barrows(1993)}]{barrows1993overview}
Howard~S Barrows. 1993.
\newblock An overview of the uses of standardized patients for teaching and evaluating clinical skills. aamc.
\newblock \emph{Academic medicine}, 68(6):443--51.

\bibitem[{Bialer et~al.(2011)Bialer, Kissane, Brown, Levin, and Bylund}]{bialer_responding_2011}
Philip~A. Bialer, David Kissane, Richard Brown, Tomer Levin, and Carma Bylund. 2011.
\newblock \href {https://doi.org/10.1017/S147895151100037X} {Responding to patient anger: {Development} and evaluation of an oncology communication skills training module}.
\newblock \emph{Palliative and Supportive Care}, 9(4):359--365.

\bibitem[{Blake and Blake(2019)}]{blake_improving_2019}
Tim Blake and Tayler Blake. 2019.
\newblock \href {https://doi.org/10.1016/j.teln.2019.06.003} {Improving {Therapeutic} {Communication} in {Nursing} {Through} {Simulation} {Exercise}}.
\newblock \emph{Teaching and Learning in Nursing}, 14(4):260--264.

\bibitem[{Brown and Bylund(2008)}]{brown_communication_2008}
Richard~F. Brown and Carma~L. Bylund. 2008.
\newblock \href {https://doi.org/10.1097/ACM.0b013e31815c631e} {Communication {Skills} {Training}: {Describing} a {New} {Conceptual} {Model}:}.
\newblock \emph{Academic Medicine}, 83(1):37--44.

\bibitem[{Cannity et~al.(2021)Cannity, Banerjee, Hichenberg, Leon-Nastasi, Howell, Coyle, Zaider, and Parker}]{cannity2021acceptability}
Kerry~M Cannity, Smita~C Banerjee, Shira Hichenberg, Angelina~D Leon-Nastasi, Frances Howell, Nessa Coyle, Talia Zaider, and Patricia~A Parker. 2021.
\newblock Acceptability and efficacy of a communication skills training for nursing students: Building empathy and discussing complex situations.
\newblock \emph{Nurse Education in Practice}, 50:102928.

\bibitem[{Chan et~al.(2023)Chan, Chen, Su, Yu, Xue, Zhang, Fu, and Liu}]{chan2023chateval}
Chi-Min Chan, Weize Chen, Yusheng Su, Jianxuan Yu, Wei Xue, Shanghang Zhang, Jie Fu, and Zhiyuan Liu. 2023.
\newblock Chateval: Towards better llm-based evaluators through multi-agent debate.
\newblock \emph{arXiv preprint arXiv:2308.07201}.

\bibitem[{Chen et~al.(2023)Chen, Wu, Zhu, Lan, Zhang, and Cui}]{chen_llm-empowered_2023}
Siyuan Chen, Mengyue Wu, Kenny~Q. Zhu, Kunyao Lan, Zhiling Zhang, and Lyuchun Cui. 2023.
\newblock \href {https://doi.org/10.48550/arXiv.2305.13614} {{LLM}-empowered {Chatbots} for {Psychiatrist} and {Patient} {Simulation}: {Application} and {Evaluation}}.
\newblock \emph{arXiv preprint}.
\newblock ArXiv:2305.13614 [cs].

\bibitem[{Chochinov et~al.(2013)Chochinov, McClement, Hack, McKeen, Rach, Gagnon, Sinclair, and Taylor-Brown}]{chochinov2013health}
Harvey~M Chochinov, Susan~E McClement, Thomas~F Hack, Nancy~A McKeen, Amanda~M Rach, Pierre Gagnon, Shane Sinclair, and Jill Taylor-Brown. 2013.
\newblock Health care provider communication: an empirical model of therapeutic effectiveness.
\newblock \emph{Cancer}, 119(9):1706--1713.

\bibitem[{Colson et~al.(1985)Colson, Allen, Coyne, Deering, Jehl, Kearns, and Spohn}]{colson1985patterns}
Donald~B Colson, Jon~G Allen, Lolafaye Coyne, David Deering, Nancy Jehl, William Kearns, and Herbert Spohn. 1985.
\newblock Patterns of staff perception of difficult patients in a long-term psychiatric hospital.
\newblock \emph{Psychiatric Services}, 36(2):168--172.

\bibitem[{De~Vries et~al.(2009)De~Vries, Bakker-Pieper, Alting~Siberg, van Gameren, and Vlug}]{de2009content}
Reinout~E De~Vries, Angelique Bakker-Pieper, Robert Alting~Siberg, Kim van Gameren, and Martijn Vlug. 2009.
\newblock The content and dimensionality of communication styles.
\newblock \emph{Communication Research}, 36(2):178--206.

\bibitem[{Dithole et~al.(2016)Dithole, Sibanda, Moleki, and Thupayagale-Tshweneagae}]{dithole2016exploring}
Kefalotse Dithole, Sambulelwe Sibanda, Mary~M Moleki, and Gloria Thupayagale-Tshweneagae. 2016.
\newblock Exploring communication challenges between nurses and mechanically ventilated patients in the intensive care unit: a structured review.
\newblock \emph{Worldviews on Evidence-Based Nursing}, 13(3):197--206.

\bibitem[{Elendu et~al.(2024)Elendu, Amaechi, Okatta, Amaechi, Elendu, Ezeh, and Elendu}]{elendu2024impact}
Chukwuka Elendu, Dependable~C Amaechi, Alexander~U Okatta, Emmanuel~C Amaechi, Tochi~C Elendu, Chiamaka~P Ezeh, and Ijeoma~D Elendu. 2024.
\newblock The impact of simulation-based training in medical education: A review.
\newblock \emph{Medicine}, 103(27):e38813.

\bibitem[{Ernstmeyer et~al.(2022)Ernstmeyer, Christman, and {Chippewa Valley Technical College}}]{ernstmeyer_nursing_2022}
Kimberly Ernstmeyer, Elizabeth Christman, and {Chippewa Valley Technical College}, editors. 2022.
\newblock \href {https://www.ncbi.nlm.nih.gov/books/NBK590036/} {\emph{Nursing: mental health and community concepts}}.
\newblock Open resources for nursing. Chippewa Valley Technical College, Eau Claire (WI).

\bibitem[{Fan et~al.(2025)Fan, Wei, Tang, Chen, Siyuan, Wei, and Huang}]{fan-etal-2025-ai}
Zhihao Fan, Lai Wei, Jialong Tang, Wei Chen, Wang Siyuan, Zhongyu Wei, and Fei Huang. 2025.
\newblock \href {https://aclanthology.org/2025.coling-main.680/} {{AI} hospital: Benchmarking large language models in a multi-agent medical interaction simulator}.
\newblock In \emph{Proceedings of the 31st International Conference on Computational Linguistics}, pages 10183--10213, Abu Dhabi, UAE. Association for Computational Linguistics.

\bibitem[{Gabriel et~al.(2024)Gabriel, Puri, Xu, Malgaroli, and Ghassemi}]{gabriel-etal-2024-ai}
Saadia Gabriel, Isha Puri, Xuhai Xu, Matteo Malgaroli, and Marzyeh Ghassemi. 2024.
\newblock \href {https://doi.org/10.18653/v1/2024.findings-emnlp.120} {Can {AI} relate: Testing large language model response for mental health support}.
\newblock In \emph{Findings of the Association for Computational Linguistics: EMNLP 2024}, pages 2206--2221, Miami, Florida, USA. Association for Computational Linguistics.

\bibitem[{Graf et~al.(2024)Graf, Sykownik, Gradl-Dietsch, and Masuch}]{graf2024towards}
Linda Graf, Philipp Sykownik, Gertraud Gradl-Dietsch, and Maic Masuch. 2024.
\newblock Towards believable and educational conversations with virtual patients.
\newblock \emph{Frontiers in Virtual Reality}, 5:1377210.

\bibitem[{Groves et~al.(1978)}]{groves1978taking}
James~E Groves et~al. 1978.
\newblock Taking care of the hateful patient.

\bibitem[{Hahn et~al.(1996)Hahn, Kroenke, Spitzer, Brody, Williams, Linzer, and DeGruy}]{hahn1996difficult}
Steven~R Hahn, Kurt Kroenke, Robert~L Spitzer, David Brody, Janet~BW Williams, Mark Linzer, and Frank~Verloin DeGruy. 1996.
\newblock The difficult patient: prevalence, psychopathology, and functional impairment.
\newblock \emph{Journal of general internal medicine}, 11:1--8.

\bibitem[{Hallett and Dickens(2017)}]{hallett_-escalation_2017}
Nutmeg Hallett and Geoffrey~L. Dickens. 2017.
\newblock \href {https://doi.org/10.1016/j.ijnurstu.2017.07.003} {De-escalation of aggressive behaviour in healthcare settings: {Concept} analysis}.
\newblock \emph{International Journal of Nursing Studies}, 75:10--20.

\bibitem[{{Hankyung}(2020)}]{hankyung2020}
{Hankyung}. 2020.
\newblock \href {https://www.hankyung.com/article/2020122107077} {Severe emotional labor among medical staff... nurses suffering from difficult patients}.
\newblock Online article.
\newblock Accessed: 2025-02-16.

\bibitem[{Hardavella et~al.(2017)Hardavella, Aamli-Gaagnat, Frille, Saad, Niculescu, and Powell}]{hardavella2017top}
Georgia Hardavella, Ane Aamli-Gaagnat, Armin Frille, Neil Saad, Alexandra Niculescu, and Pippa Powell. 2017.
\newblock Top tips to deal with challenging situations: doctor--patient interactions.
\newblock \emph{Breathe}, 13(2):129--135.

\bibitem[{Hicke et~al.(2025)Hicke, Geathers, Rajashekar, Chan, Jack, Sewell, Preston, Cornes, Shung, and Kizilcec}]{hicke2025medsimaisimulationformativefeedback}
Yann Hicke, Jadon Geathers, Niroop Rajashekar, Colleen Chan, Anyanate~Gwendolyne Jack, Justin Sewell, Mackenzi Preston, Susannah Cornes, Dennis Shung, and Rene Kizilcec. 2025.
\newblock \href {https://arxiv.org/abs/2503.05793} {Medsimai: Simulation and formative feedback generation to enhance deliberate practice in medical education}.
\newblock \emph{Preprint}, arXiv:2503.05793.

\bibitem[{Hinchey and Jackson(2011)}]{hinchey2011cohort}
Sherri~A Hinchey and Jeffrey~L Jackson. 2011.
\newblock A cohort study assessing difficult patient encounters in a walk-in primary care clinic, predictors and outcomes.
\newblock \emph{Journal of general internal medicine}, 26:588--594.

\bibitem[{Ho et~al.(2021)Ho, Stenhouse, and Snowden}]{ho2021quite}
Szu-Szu Ho, Rosie Stenhouse, and Austyn Snowden. 2021.
\newblock ‘it was quite a shock’: A qualitative study of the impact of organisational and personal factors on newly qualified nurses' experiences.
\newblock \emph{Journal of Clinical Nursing}, 30(15-16):2373--2385.

\bibitem[{Holderried et~al.(2024{\natexlab{a}})Holderried, Stegemann-Philipps, Herrmann-Werner, Festl-Wietek, Holderried, Eickhoff, and Mahling}]{info:doi/10.2196/59213}
Friederike Holderried, Christian Stegemann-Philipps, Anne Herrmann-Werner, Teresa Festl-Wietek, Martin Holderried, Carsten Eickhoff, and Moritz Mahling. 2024{\natexlab{a}}.
\newblock \href {https://doi.org/10.2196/59213} {A language model--powered simulated patient with automated feedback for history taking: Prospective study}.
\newblock \emph{JMIR Med Educ}, 10:e59213.

\bibitem[{Holderried et~al.(2024{\natexlab{b}})Holderried, Stegemann-Philipps, Herschbach, Moldt, Nevins, Griewatz, Holderried, Herrmann-Werner, Festl-Wietek, Mahling et~al.}]{holderried2024generative}
Friederike Holderried, Christian Stegemann-Philipps, Lea Herschbach, Julia-Astrid Moldt, Andrew Nevins, Jan Griewatz, Martin Holderried, Anne Herrmann-Werner, Teresa Festl-Wietek, Moritz Mahling, et~al. 2024{\natexlab{b}}.
\newblock A generative pretrained transformer (gpt)--powered chatbot as a simulated patient to practice history taking: Prospective, mixed methods study.
\newblock \emph{JMIR medical education}, 10(1):e53961.

\bibitem[{Howick et~al.(2018)Howick, Moscrop, Mebius, Fanshawe, Lewith, Bishop, Mistiaen, Roberts, Dieninytė, Hu, Aveyard, and Onakpoya}]{howick_effects_2018}
Jeremy Howick, Andrew Moscrop, Alexander Mebius, Thomas~R. Fanshawe, George Lewith, Felicity~L. Bishop, Patriek Mistiaen, Nia~W. Roberts, Eglė Dieninytė, Xiao-Yang Hu, Paul Aveyard, and Igho~J. Onakpoya. 2018.
\newblock \href {https://doi.org/10.1177/0141076818769477} {Effects of empathic and positive communication in healthcare consultations: a systematic review and meta-analysis}.
\newblock \emph{Journal of the Royal Society of Medicine}, 111(7):240--252.

\bibitem[{Hoyt and Kerns(1999)}]{hoyt1999magnitude}
William~T Hoyt and Michael-David Kerns. 1999.
\newblock Magnitude and moderators of bias in observer ratings: A meta-analysis.
\newblock \emph{Psychological Methods}, 4(4):403.

\bibitem[{{INACSL}(2023)}]{INACSL2023}
{INACSL}. 2023.
\newblock \href {https://www.uthsc.edu/simulation/resources/documents/sim-template-sp.doc} {Healthcare simulation standards of best practice: Simulation design}.
\newblock Accessed: 2025-02-15.

\bibitem[{Jang and Jung(2024)}]{jang_evaluating_2024}
Woori Jang and Seohyon Jung. 2024.
\newblock \href {https://doi.org/10.18653/v1/2024.nlp4dh-1.34} {Evaluating {LLM} {Performance} in {Character} {Analysis}: {A} {Study} of {Artificial} {Beings} in {Recent} {Korean} {Science} {Fiction}}.
\newblock In \emph{Proceedings of the 4th {International} {Conference} on {Natural} {Language} {Processing} for {Digital} {Humanities}}, pages 339--351, Miami, USA. Association for Computational Linguistics.

\bibitem[{Jiang et~al.(2025)Jiang, Black, Geng, Park, Ng, and Chen}]{jiang_medagentbench_2025}
Yixing Jiang, Kameron~C. Black, Gloria Geng, Danny Park, Andrew~Y. Ng, and Jonathan~H. Chen. 2025.
\newblock \href {https://doi.org/10.48550/arXiv.2501.14654} {{MedAgentBench}: {Dataset} for {Benchmarking} {LLMs} as {Agents} in {Medical} {Applications}}.
\newblock \emph{arXiv preprint}.
\newblock ArXiv:2501.14654 [cs].

\bibitem[{Kardong-Edgren et~al.(2024)Kardong-Edgren, Wells-Beede, and Wands}]{kardong-edgren_student_2024}
Suzie Kardong-Edgren, Elizabeth Wells-Beede, and LisaMarie Wands. 2024.
\newblock \href {https://doi.org/10.1016/j.teln.2024.02.007} {Student abuse in simulation: causes and solutions}.
\newblock \emph{Teaching and Learning in Nursing}, 19(2):145--149.

\bibitem[{Kim et~al.(2020)Kim, Jung, Kim, and You}]{kim2020factors}
Eun~Gyung Kim, Myun~Sook Jung, Jong~Kyung Kim, and Sun~Ju You. 2020.
\newblock Factors affecting new graduate nurses' intention on retention in hospitals: Focused on nursing organizational culture, empowering leadership and organizational socialization.
\newblock \emph{Journal of Korean Academy of Nursing Administration}, 26(1):31--41.

\bibitem[{Kim et~al.(2024)Kim, Lee, Kim, and Kim}]{kim_development_2024}
Jundong Kim, Hye-Yoon Lee, Ji-Hwan Kim, and Chang-Eop Kim. 2024.
\newblock \href {https://doi.org/10.13048/jkm.24067} {Development of an {LLM}-based {CPX} {Practicing} {Chatbot} for {Korean} {Medicine} {Education}: {Implementation} of {Automated} {Scoring} and {Feedback} {Generation} {Framework}}.
\newblock \emph{Journal of Korean Medicine}, 45(4):215--230.

\bibitem[{Kits(1990)}]{kits1990nurses}
Robert~J Kits. 1990.
\newblock Nurses and unpopular patients.
\newblock \emph{AJN The American Journal of Nursing}, 90(6):62--68.

\bibitem[{Kleinsmith et~al.(2015)Kleinsmith, Rivera-Gutierrez, Finney, Cendan, and Lok}]{kleinsmith_understanding_2015}
Andrea Kleinsmith, Diego Rivera-Gutierrez, Glen Finney, Juan Cendan, and Benjamin Lok. 2015.
\newblock \href {https://doi.org/10.1016/j.chb.2015.05.033} {Understanding empathy training with virtual patients}.
\newblock \emph{Computers in Human Behavior}, 52:151--158.

\bibitem[{{KN News}(2016)}]{knnews2016}
{KN News}. 2016.
\newblock \href {https://www.knnews.co.kr/news/articleView.php?idxno=1176332} {Nurses face severe emotional labor… struggling to handle 'difficult patients'}.
\newblock Online article.
\newblock Accessed: 2025-02-16.

\bibitem[{Koo et~al.(2024)Koo, Lee, Raheja, Park, Kim, and Kang}]{koo_benchmarking_2024}
Ryan Koo, Minhwa Lee, Vipul Raheja, Jong~Inn Park, Zae~Myung Kim, and Dongyeop Kang. 2024.
\newblock \href {https://doi.org/10.48550/arXiv.2309.17012} {Benchmarking {Cognitive} {Biases} in {Large} {Language} {Models} as {Evaluators}}.
\newblock \emph{arXiv preprint}.
\newblock ArXiv:2309.17012 [cs].

\bibitem[{Kraut et~al.(1982)Kraut, Lewis, and Swezey}]{kraut_listener_1982}
Robert~E. Kraut, Steven~H. Lewis, and Lawrence~W. Swezey. 1982.
\newblock \href {https://doi.org/10.1037/0022-3514.43.4.718} {Listener responsiveness and the coordination of conversation.}
\newblock \emph{Journal of Personality and Social Psychology}, 43(4):718--731.

\bibitem[{Li et~al.(2024)Li, Zeng, Zhong, Zhang, Zhang, and Zou}]{li_leveraging_2024}
Yanzeng Li, Cheng Zeng, Jialun Zhong, Ruoyu Zhang, Minhao Zhang, and Lei Zou. 2024.
\newblock \href {https://doi.org/10.48550/arXiv.2404.13066} {Leveraging {Large} {Language} {Model} as {Simulated} {Patients} for {Clinical} {Education}}.
\newblock \emph{arXiv preprint}.
\newblock ArXiv:2404.13066 [cs].

\bibitem[{Liu et~al.(2024)Liu, Moosavi, and Lin}]{liu_llms_2024}
Yiqi Liu, Nafise~Sadat Moosavi, and Chenghua Lin. 2024.
\newblock \href {https://doi.org/10.48550/arXiv.2311.09766} {{LLMs} as {Narcissistic} {Evaluators}: {When} {Ego} {Inflates} {Evaluation} {Scores}}.
\newblock \emph{arXiv preprint}.
\newblock ArXiv:2311.09766 [cs].

\bibitem[{Liusie et~al.(2024)Liusie, Manakul, and Gales}]{liusie_llm_2024}
Adian Liusie, Potsawee Manakul, and Mark J.~F. Gales. 2024.
\newblock \href {https://doi.org/10.48550/arXiv.2307.07889} {{LLM} {Comparative} {Assessment}: {Zero}-shot {NLG} {Evaluation} through {Pairwise} {Comparisons} using {Large} {Language} {Models}}.
\newblock \emph{arXiv preprint}.
\newblock ArXiv:2307.07889 [cs].

\bibitem[{Louie et~al.(2024)Louie, Nandi, Fang, Chang, Brunskill, and Yang}]{louie-etal-2024-roleplay}
Ryan Louie, Ananjan Nandi, William Fang, Cheng Chang, Emma Brunskill, and Diyi Yang. 2024.
\newblock \href {https://doi.org/10.18653/v1/2024.emnlp-main.591} {Roleplay-doh: Enabling domain-experts to create {LLM}-simulated patients via eliciting and adhering to principles}.
\newblock In \emph{Proceedings of the 2024 Conference on Empirical Methods in Natural Language Processing}, pages 10570--10603, Miami, Florida, USA. Association for Computational Linguistics.

\bibitem[{MacLean et~al.(2017)MacLean, Kelly, Geddes, and Della}]{maclean2017use}
Sharon MacLean, Michelle Kelly, Fiona Geddes, and Phillip Della. 2017.
\newblock Use of simulated patients to develop communication skills in nursing education: An integrative review.
\newblock \emph{Nurse education today}, 48:90--98.

\bibitem[{Madula et~al.(2018)Madula, Kalembo, Yu, and Kaminga}]{madula_healthcare_2018}
Precious Madula, Fatch~Welcome Kalembo, Hong Yu, and Atipatsa~Chiwanda Kaminga. 2018.
\newblock \href {https://doi.org/10.1186/s12978-018-0580-x} {Healthcare provider-patient communication: a qualitative study of women’s perceptions during childbirth}.
\newblock \emph{Reproductive Health}, 15(1):135.

\bibitem[{Marcum(2015)}]{marcum2015caring}
James~A Marcum. 2015.
\newblock Caring for patients during challenging clinical encounters.
\newblock \emph{Journal of Evaluation in Clinical Practice}, 21(3):404--409.

\bibitem[{McGaghie et~al.(2010)McGaghie, Issenberg, Petrusa, and Scalese}]{mcgaghie2010critical}
William~C McGaghie, S~Barry Issenberg, Emil~R Petrusa, and Ross~J Scalese. 2010.
\newblock A critical review of simulation-based medical education research: 2003--2009.
\newblock \emph{Medical education}, 44(1):50--63.

\bibitem[{Nestel and Bearman(2014)}]{nestel2014simulated}
Debra Nestel and Margaret Bearman. 2014.
\newblock \emph{Simulated patient methodology: theory, evidence and practice}.
\newblock John Wiley \& Sons.

\bibitem[{Pascucci et~al.(2014)Pascucci, Weinstock, O’Connor, Fancy, and Meyer}]{pascucci_integrating_2014}
Robert~C. Pascucci, Peter~H. Weinstock, Brigid~E. O’Connor, Kristina~M. Fancy, and Elaine~C. Meyer. 2014.
\newblock \href {https://doi.org/10.1097/SIH.0b013e3182a3ded7} {Integrating {Actors} {Into} a {Simulation} {Program}: {A} {Primer}}.
\newblock \emph{Simulation in Healthcare: The Journal of the Society for Simulation in Healthcare}, 9(2):120--126.

\bibitem[{Patak et~al.(2009)Patak, Wilson-Stronks, Costello, Kleinpell, Henneman, Person, and Happ}]{patak2009improving}
Lance Patak, Amy Wilson-Stronks, John Costello, Ruth~M Kleinpell, Elizabeth~A Henneman, Colleen Person, and Mary~Beth Happ. 2009.
\newblock Improving patient-provider communication: a call to action.
\newblock \emph{JONA: The Journal of Nursing Administration}, 39(9):372--376.

\bibitem[{Peimani et~al.(2020)Peimani, Nasli-Esfahani, and Sadeghi}]{peimani2020patients}
Maryam Peimani, Ensieh Nasli-Esfahani, and Roya Sadeghi. 2020.
\newblock Patients’ perceptions of patient--provider communication and diabetes care: A systematic review of quantitative and qualitative studies.
\newblock \emph{Chronic illness}, 16(1):3--22.

\bibitem[{Pines et~al.(2021)Pines, Giles, and Watson}]{Pines20216281}
Rachyl Pines, Howard Giles, and Bernadette Watson. 2021.
\newblock \href {https://doi.org/10.2478/plc-2021-0004} {Managing patient aggression in healthcare: Initial testing of a communication accommodation theory intervention}.
\newblock \emph{Psychology of Language and Communication}, 25(1):62--81.

\bibitem[{Podsakoff et~al.(2003)Podsakoff, MacKenzie, Lee, and Podsakoff}]{podsakoff2003common}
Philip~M Podsakoff, Scott~B MacKenzie, Jeong-Yeon Lee, and Nathan~P Podsakoff. 2003.
\newblock Common method biases in behavioral research: a critical review of the literature and recommended remedies.
\newblock \emph{Journal of applied psychology}, 88(5):879.

\bibitem[{Price et~al.(2024)Price, Armitage, Bee, Brooks, Lovell, Butler, Cree, Fishwick, Grundy, Johnston, Mcpherson, Riches, Scott, Walker, and Papastavrou~Brooks}]{price_-escalating_2024}
Owen Price, Christopher~J. Armitage, Penny Bee, Helen Brooks, Karina Lovell, Debbie Butler, Lindsey Cree, Paul Fishwick, Andrew Grundy, Isobel Johnston, Peter Mcpherson, Holly Riches, Anne Scott, Lauren Walker, and Cat Papastavrou~Brooks. 2024.
\newblock \href {https://doi.org/10.1186/s12888-024-05920-y} {De-escalating aggression in acute inpatient mental health settings: a behaviour change theory-informed, secondary qualitative analysis of staff and patient perspectives}.
\newblock \emph{BMC Psychiatry}, 24(1):548.

\bibitem[{Richmond et~al.(2012)Richmond, Berlin, Fishkind, Holloman, Zeller, Wilson, Rifai, and Ng}]{richmond_verbal_2012}
Janet Richmond, Jon Berlin, Avrim Fishkind, Garland Holloman, Scott Zeller, Michael Wilson, Muhamad~Aly Rifai, and Anthony Ng. 2012.
\newblock \href {https://doi.org/10.5811/westjem.2011.9.6864} {Verbal {De}-escalation of the {Agitated} {Patient}: {Consensus} {Statement} of the {American} {Association} for {Emergency} {Psychiatry} {Project} {BETA} {De}-escalation {Workgroup}}.
\newblock \emph{Western Journal of Emergency Medicine}, 13(1):17--25.

\bibitem[{Sardesai et~al.(2024)Sardesai, Russo, Martin, and Sardesai}]{sardesai2024utilizing}
Neil Sardesai, Paolo Russo, Jonathan Martin, and Anand Sardesai. 2024.
\newblock Utilizing generative conversational artificial intelligence to create simulated patient encounters: a pilot study for anaesthesia training.
\newblock \emph{Postgraduate Medical Journal}, 100(1182):237--241.

\bibitem[{Schmidgall et~al.(2024)Schmidgall, Ziaei, Harris, Reis, Jopling, and Moor}]{schmidgall_agentclinic_2024}
Samuel Schmidgall, Rojin Ziaei, Carl Harris, Eduardo Reis, Jeffrey Jopling, and Michael Moor. 2024.
\newblock \href {https://doi.org/10.48550/arXiv.2405.07960} {{AgentClinic}: a multimodal agent benchmark to evaluate {AI} in simulated clinical environments}.
\newblock \emph{arXiv preprint}.
\newblock ArXiv:2405.07960 [cs].

\bibitem[{Serour et~al.(2010)Serour, Othman, and Khalifah}]{serour2010difficult}
Maleka Serour, Heyam~Al Othman, and Ghada~Al Khalifah. 2010.
\newblock Difficult patients or difficult doctors: an analysis of problematic consultations.

\bibitem[{Sheldon(2009)}]{sheldon_communication_2009}
Lisa~Kennedy Sheldon. 2009.
\newblock \emph{Communication for nurses: talking with patients}, 2nd ed edition.
\newblock Jones and Bartlett Publishers, Sudbury, Mass.
\newblock OCLC: 243818592.

\bibitem[{Son et~al.(2017)Son, Lee, and Cho}]{son2017affecting}
Haeng-Mi Son, Eun~Hee Lee, and Kyung~Sook Cho. 2017.
\newblock Affecting factors of new nurse's intention to retention in hospitals.
\newblock \emph{Journal of muscle and joint health}, 24(3):205--216.

\bibitem[{Spencer et~al.(2018)Spencer, Johnson, and Smith}]{spencer_-escalation_2018}
Sally Spencer, Paula Johnson, and Ian~C Smith. 2018.
\newblock \href {https://doi.org/10.1002/14651858.CD012034.pub2} {De-escalation techniques for managing non-psychosis induced aggression in adults}.
\newblock \emph{Cochrane Database of Systematic Reviews}, 2018(7).

\bibitem[{Steenstra et~al.(2025)Steenstra, Nouraei, and Bickmore}]{10.1145/3706598.3714014}
Ian Steenstra, Farnaz Nouraei, and Timothy Bickmore. 2025.
\newblock \href {https://doi.org/10.1145/3706598.3714014} {Scaffolding empathy: Training counselors with simulated patients and utterance-level performance visualizations}.
\newblock In \emph{Proceedings of the 2025 CHI Conference on Human Factors in Computing Systems}, CHI '25, New York, NY, USA. Association for Computing Machinery.

\bibitem[{Stein et~al.(2022)Stein, Cannity, Weiner, Hichenberg, Leon-Nastasi, Banerjee, and Parker}]{stein2022general}
Deborah Stein, Kerry Cannity, Richard Weiner, Shira Hichenberg, Angelina Leon-Nastasi, Smita Banerjee, and Patricia Parker. 2022.
\newblock General and unique communication skills challenges for advanced practice providers: a mixed-methods study.
\newblock \emph{Journal of the Advanced Practitioner in Oncology}, 13(1):32.

\bibitem[{Stephen et~al.(2020)Stephen, Kostovich, and O'Rourke}]{stephen_psychological_2020}
Lee-Anne Stephen, Carol Kostovich, and Jenny O'Rourke. 2020.
\newblock \href {https://doi.org/10.1016/j.ecns.2020.06.010} {Psychological {Safety} in {Simulation}: {Prelicensure} {Nursing} {Students}’ {Perceptions}}.
\newblock \emph{Clinical Simulation in Nursing}, 47:25--31.

\bibitem[{Tanana et~al.(2019)Tanana, Soma, Srikumar, Atkins, and Imel}]{tanana_development_2019}
Michael~J Tanana, Christina~S Soma, Vivek Srikumar, David~C Atkins, and Zac~E Imel. 2019.
\newblock \href {https://doi.org/10.2196/12529} {Development and {Evaluation} of {ClientBot}: {Patient}-{Like} {Conversational} {Agent} to {Train} {Basic} {Counseling} {Skills}}.
\newblock \emph{Journal of Medical Internet Research}, 21(7):e12529.

\bibitem[{TMLT(2022{\natexlab{a}})}]{TMLT1}
Texas Medical Liability~Trust TMLT. 2022{\natexlab{a}}.
\newblock \href {https://www.youtube.com/watch?v=HJi_AAvb3uA} {De-escalation video 1: How to identify and prepare to meet with a disruptive patient}.

\bibitem[{TMLT(2022{\natexlab{b}})}]{TMLT3}
Texas Medical Liability~Trust TMLT. 2022{\natexlab{b}}.
\newblock \href {https://www.youtube.com/watch?v=cbQUrfYxNtg} {De-escalation video 3: Actions to avoid...and actions to take}.

\bibitem[{Townsley et~al.(2023)Townsley, Li-Wang, and Katta}]{townsley2023patient}
Alexandra Townsley, Jennifer Li-Wang, and Rajani Katta. 2023.
\newblock When patient rudeness impacts care: a review of incivility in healthcare.
\newblock \emph{Cureus}, 15(6).

\bibitem[{Wallace et~al.(2002)Wallace, Rao, and Haslam}]{wallace2002simulated}
Jeremy Wallace, Ranga Rao, and Richard Haslam. 2002.
\newblock Simulated patients and objective structured clinical examinations: review of their use in medical education.
\newblock \emph{Advances in Psychiatric treatment}, 8(5):342--348.

\bibitem[{Wang et~al.(2024{\natexlab{a}})Wang, Yao, Yang, Zhou, Li, Wang, Xu, and Yu}]{wang2024notechat}
Junda Wang, Zonghai Yao, Zhichao Yang, Huixue Zhou, Rumeng Li, Xun Wang, Yucheng Xu, and Hong Yu. 2024{\natexlab{a}}.
\newblock Notechat: a dataset of synthetic patient-physician conversations conditioned on clinical notes.
\newblock In \emph{Findings of the Association for Computational Linguistics ACL 2024}, pages 15183--15201.

\bibitem[{Wang et~al.(2023)Wang, Li, Chen, Cai, Zhu, Lin, Cao, Liu, Liu, and Sui}]{wang_large_2023}
Peiyi Wang, Lei Li, Liang Chen, Zefan Cai, Dawei Zhu, Binghuai Lin, Yunbo Cao, Qi~Liu, Tianyu Liu, and Zhifang Sui. 2023.
\newblock \href {https://doi.org/10.48550/arXiv.2305.17926} {Large {Language} {Models} are not {Fair} {Evaluators}}.
\newblock \emph{arXiv preprint}.
\newblock ArXiv:2305.17926 [cs].

\bibitem[{Wang et~al.(2024{\natexlab{b}})Wang, Milani, Chiu, Zhi, Eack, Labrum, Murphy, Jones, Hardy, Shen, Fang, and Chen}]{wang-etal-2024-patient}
Ruiyi Wang, Stephanie Milani, Jamie~C. Chiu, Jiayin Zhi, Shaun~M. Eack, Travis Labrum, Samuel~M Murphy, Nev Jones, Kate~V Hardy, Hong Shen, Fei Fang, and Zhiyu Chen. 2024{\natexlab{b}}.
\newblock \href {https://doi.org/10.18653/v1/2024.emnlp-main.711} {{PATIENT}-$\psi$: Using large language models to simulate patients for training mental health professionals}.
\newblock In \emph{Proceedings of the 2024 Conference on Empirical Methods in Natural Language Processing}, pages 12772--12797, Miami, Florida, USA. Association for Computational Linguistics.

\bibitem[{Wind et~al.(2004)Wind, Van~Dalen, Muijtjens, and Rethans}]{wind2004assessing}
Lidewij~A Wind, Jan Van~Dalen, Arno~MM Muijtjens, and Jan-Joost Rethans. 2004.
\newblock Assessing simulated patients in an educational setting: the masp (maastricht assessment of simulated patients).
\newblock \emph{Medical Education}, 38(1):39--44.

\bibitem[{Wu et~al.(2023)Wu, Gong, Shou, Liang, and Jiang}]{wu_large_2023}
Ning Wu, Ming Gong, Linjun Shou, Shining Liang, and Daxin Jiang. 2023.
\newblock \href {https://doi.org/10.48550/ARXIV.2303.15078} {Large {Language} {Models} are {Diverse} {Role}-{Players} for {Summarization} {Evaluation}}.
\newblock \emph{arXiv preprint}.
\newblock Version Number: 3.

\bibitem[{{Yonhap News}(2024)}]{yonhap2024}
{Yonhap News}. 2024.
\newblock \href {https://www.yna.co.kr/view/AKR20240319155800530} {{More than half of new nurses quit within a year... "Excessive workload and maladaptation"}}.
\newblock Accessed: 2025-02-15.

\bibitem[{Zhang et~al.(2023)Zhang, Yu, Yu, Lv, Liu, Huang, Xu, and Li}]{zhang_wider_2023}
Xinghua Zhang, Bowen Yu, Haiyang Yu, Yangyu Lv, Tingwen Liu, Fei Huang, Hongbo Xu, and Yongbin Li. 2023.
\newblock \href {https://doi.org/10.48550/arXiv.2308.01862} {Wider and {Deeper} {LLM} {Networks} are {Fairer} {LLM} {Evaluators}}.
\newblock \emph{arXiv preprint}.
\newblock ArXiv:2308.01862 [cs].

\bibitem[{Zhou et~al.(2024)Zhou, Pang, Shen, and Cheng}]{zhou-etal-2024-think}
Junkai Zhou, Liang Pang, Huawei Shen, and Xueqi Cheng. 2024.
\newblock \href {https://doi.org/10.18653/v1/2024.findings-naacl.248} {Think before you speak: Cultivating communication skills of large language models via inner monologue}.
\newblock In \emph{Findings of the Association for Computational Linguistics: NAACL 2024}, pages 3925--3951, Mexico City, Mexico. Association for Computational Linguistics.

\bibitem[{Zhu et~al.(2024)Zhu, Wang, Zhao, Xu, and Xie}]{zhu_dynamic_2024}
Kaijie Zhu, Jindong Wang, Qinlin Zhao, Ruochen Xu, and Xing Xie. 2024.
\newblock \href {https://doi.org/10.48550/ARXIV.2402.14865} {Dynamic {Evaluation} of {Large} {Language} {Models} by {Meta} {Probing} {Agents}}.
\newblock \emph{arXiv preprint}.
\newblock Version Number: 2.

\bibitem[{Zhu et~al.(2023)Zhu, Wang, and Wang}]{zhu_judgelm_2023}
Lianghui Zhu, Xinggang Wang, and Xinlong Wang. 2023.
\newblock \href {https://doi.org/10.48550/arXiv.2310.17631} {{JudgeLM}: {Fine}-tuned {Large} {Language} {Models} are {Scalable} {Judges}}.
\newblock \emph{arXiv preprint}.
\newblock ArXiv:2310.17631 [cs].

\bibitem[{Ziv et~al.(2006)Ziv, Wolpe, Small, and Glick}]{ziv2006simulation}
Amitai Ziv, Paul~Root Wolpe, Stephen~D Small, and Shimon Glick. 2006.
\newblock Simulation-based medical education: an ethical imperative.
\newblock \emph{Simulation in Healthcare}, 1(4):252--256.

\end{thebibliography}

\onecolumn
\appendix

\section{VP Case Development Pipeline}
\label{app:VP_Case_Development_Prompt}

\subsection{4 Types of Challenging Patients (\citealt{groves1978taking})}
\label{app:VP-patient_type}

\begin{table*}[h]
\centering
\fontsize{9pt}{12pt}\selectfont
\resizebox{\textwidth}{!}{%
\begin{tabular}{@{}lp{13cm}@{}}
\toprule
\textbf{Type} & \textbf{Description} \\
\midrule
Overdependent & Relies heavily on nurses to alleviate anxiety about illness \\
 & Frequently calls nurses or seeks reassurance for every concern \\ 
\midrule
Authoritative & Attempts to exploit healthcare providers through intimidation or guilt \\
 & Believes excessive anger and unreasonable demands are justified as a defense mechanism \\
\midrule
Aggressive & Openly displays anger and hostility \\
 & Threatens or resorts to violent behavior toward nurses \\
\midrule
Uncooperative & Remains overly pessimistic about treatment or actively impedes care \\
 & Sometimes displays dependent behaviors while denying the possibility of recovery \\
\bottomrule
\end{tabular}%
}
\caption{Types of Challenging Patient Interactions}
\label{tab:patient_types}
\end{table*}

\subsection{Incorporating Literature}
\label{app:VP-Training_Context_Prompt}

\texttt{
 \\
The following describes types of problematic patients who create difficulties in nurse-patient communication:\\
\\
Type 1: Overly Dependent Patients
\begin{itemize}
    \item[] Attempt to resolve psychological fears about their illness by becoming excessively dependent on nurses
    \item[] Individuals who worry excessively about their illness or use the nurse call button too frequently
\end{itemize}
Type 2: Overly Authoritative Patients
\begin{itemize}
    \item[] Attempt to manipulate doctors through threats or inducing guilt
    \item[] Feel threatened by nurses having power over their life and death, and thus believe they have the right to express anger and make excessive or inappropriate demands as a defense mechanism
    \item[] Patients who never praise or thank nurses, are obsessed with filing lawsuits against nurses, or constantly complain
\end{itemize}
Type 3: Threatening and Violent Patients
\begin{itemize}
    \item[] Patients who argue with others
    \item[] Patients who express anger and hostility
    \item[] Patients who are violent towards their family members or objects
    \item[] Patients who threaten or show violent behavior towards nurses
\end{itemize}
Type 4: Non-compliant Patients
\begin{itemize}
    \item[] Patients who are excessively pessimistic about treatment or directly engage in behaviors that interfere with treatment
    \item[] Continued smoking by lung cancer patients
    \item[] Continued drinking by alcoholic patients
\end{itemize}
While overly dependent, they use defense mechanisms to deny their chances of survival\\
Give up hope of being treated and take pride in self-destructive behaviors\\
Actually derive satisfaction from interfering with treatment\\
\\
Background: We plan to conduct simulation training for nurse-patient communication for new nurses.\\
Task: For each of the four patient types above, describe 5 specific situations in Korean where these patients create problems in nurse-patient communication in Korean medical and surgical general wards.\\
}

\subsection{Basic Profile Generation}
\label{app:VP-Basic_Profile_Generation_Prompt}
\texttt{
 \\
Create highly detailed, realistic, and vivid patient profiles for the 5 situations of the specified type. Focus on making the 5 patients as diverse as possible while keeping them realistic for Korean ward settings.
\begin{itemize}
    \item[]<patient\_type\_description>
    \begin{itemize}
        \item[]\{PATIENT\_TYPE\_DESCRIPTION\}
    \end{itemize}
    \item[]</patient\_type\_description>
    \item[]<rule> 
    \begin{itemize}
        \item Write in JSON format
        \item Keys should be in English, Values should be in Korean
        \item Add "type" as a key to all patient profiles with a value of 1
    \end{itemize}
    \item[]</rule>
\end{itemize}
For each patient profile, include the following information:
\begin{itemize}
    \item[-] Brief description of client
    \item[-]name
    \item[-]gender
    \item[-]age
    \item[-]religion
    \item[-]height
    \item[-]weight
    \item[-]Chief complaint: 1-2 Quotes
    \item[-]History of present illness
    \item[-]social history
    \item[-]past medical history
    \item[-]past surgical history \& date
    \item[-]family medical history
    \item[-]allergies
    \item[-]immunization
    \item[-]medication
    \item[-]primary diagnosis
    \item[-]communication style
\end{itemize}
}

\subsection{Communication Traits Generation}
\label{app:VP-Communication_Traits_Generation}

\texttt{
 \\
You are tasked with analyzing a virtual patient's profile for a nursing communication training simulation. The patient profile is provided below:
\begin{itemize}
    \item[]<patient\_profile> \{PATIENT\_PROFILE\} </patient\_profile>
\end{itemize}
Please follow these steps:
\begin{itemize}
    \item[1.]Carefully review the patient profile, paying special attention to the communication style section.
    \item[2.]Summarize the patient's communication style in five sentences or fewer in Korean. Focus on the key characteristics that define how this patient interacts with healthcare providers.
    \item[3.]Within your summary, make sure to highlight the problematic aspects of the patient's communication style. These are behaviors or tendencies that may pose challenges for nurses during interactions.
    \item[4.]Do not specify a numerical scale. Instead, focus on describing how these communication characteristics manifest in actual conversations.
    \item[5.]Write two example expressions that the patient might use when first addressing a passing nurse. The expressions should clearly convey the patient's needs. These should be direct quotes that illustrate the patient's communication style. Make sure to write realistic and natural Korean expressions that the patient would likely use in real-life situations. Avoid overly dramatic expressions.
    \item[6.]Focus on portraying a realistic patient image for this research-based simulation. However, exclude any content about complaining to higher authorities such as the hospital director or head nurse.
    \item[7.]Present your analysis in the following format:
    \begin{itemize}
        \item[]<analysis>
        \begin{itemize}
            \item[]<summary>
            \begin{itemize}
                \item[]\{Your five-sentence summary of the patient's communication style, including problematic aspects\}
            \end{itemize}
            \item[]</summary>
            \item[]<example\_expressions>
            \begin{itemize}
                \item[1.]"[First example expression]"
                \item[2.]"[Second example expression]"
            \end{itemize}
            \item[]</example\_expressions>
        \end{itemize}
        \item[]</analysis
    \end{itemize}
\end{itemize}
Remember to base your analysis solely on the information provided in the patient profile. Do not invent or assume details that are not present in the given information.
}

\section{Generated VP Cases}
\label{app:Generated_Patient_Profiles}

The following are the example cases of 4 different types of virtual patient agents.\\

\subsection{Type 1: Overdependent}
\label{app:Profile-overdependent}

\begin{table}[H]
\centering
\renewcommand{\arraystretch}{1.2}
\resizebox{\textwidth}{!}{
\begin{tabular}{m{4.5cm} m{\textwidth}}
\toprule
\textbf{Information} & \textbf{Prompt} \\
\midrule
ID & 0\\
\midrule
Type & overdependent\\
\midrule
Name & Lee Mikyung\\
\midrule
Situation &  A highly dependent patient repeatedly calls for nurses during night shifts expressing anxiety. Despite it being sleep time, they press the call button every 30 minutes, constantly demand attention and conversation even when nurses need to focus on other patients. Whenever nurses try to leave after completing their tasks, the patient becomes anxious and repeatedly asks them to stay.\\
\midrule
Chief Complaint & "(with tearful voice) My heart keeps pounding and I can’t sleep... I’m not getting worse, am I? Could you please stay and talk with me for a while?"\\
\midrule
Gender & Female\\
\midrule
Age & 55\\
\midrule
Religion & None\\
\midrule
Height & 162cm\\
\midrule
Weight & 58kg\\
\midrule
Main Symptom & Anxiety and insomnia due to breast cancer\\
\midrule
History of Present Illness & Breast mass discovered 2 weeks ago, diagnosed as malignant after biopsy, awaiting surgery\\
\midrule
Social History & Freelance designer, divorced, no children\\
\midrule
Past Medical History & Depression (5 years)\\
\midrule
Past Surgical History \& Date & Appendectomy (20 years ago)\\
\midrule
Family Medical History & Sister - Breast cancer\\
\midrule
Allergies & Latex\\
\midrule
Immunization & Shingles vaccine (1 year ago)\\
\midrule
Medication & Escitalopram 10mg qd\\
\midrule
Primary Diagnosis & Stage 2 right breast cancer\\
\midrule
Communication Summary & The patient shows highly dependent tendencies, strongly requiring continuous interaction and emotional support from medical staff. Due to high anxiety, they repeatedly express the same concerns and constantly seek attention and reassurance from nurses. While scattered and inconsistent in conversation, they maintain a very polite and compliant attitude. Their frequent night-time calls and excessive demands for staff time and attention can interfere with care for other patients. They show high levels of anxiety and worry, consistently trying to express their symptoms and discomfort.\\
\midrule
First Statement & "Could we talk for a moment while you’re passing by? My chest feels tight, could you please stay with me a little longer?"\\
\bottomrule
\end{tabular}}
\caption{Generated case example of the overdependent VP}
\label{tab:profile-example1}
\end{table}

\vspace{5mm}
\subsection{Type 2: Overly Authoritative}
\label{app:Profile-authoritative}

\noindent
\raggedbottom

\begin{table}[H]
\centering
\renewcommand{\arraystretch}{1.2}
\resizebox{\textwidth}{!}{
\begin{tabular}{m{4.5cm} m{\textwidth}}
\toprule
\textbf{Information} & \textbf{Prompt} \\
\midrule
ID & 2\\
\midrule
Type & overly authoriative\\
\midrule
Name & Park Jeongho\\
\midrule
Situation &  An authoritative patient, assigned to a four-bed ward due to unavailability of private rooms, expresses excessive complaints about room temperature, noise, meal quality, and cleaning conditions, demanding immediate action. Mentioning their connections with corporate executives and familiarity with the hospital director, they demand special treatment and disparage nurses' professionalism and service quality. Without considering other patients' comfort, they demand room environment improvements based solely on their standards and threaten to complain to the head nurse if their demands aren't immediately met.\\
\midrule
Chief Complaint & "Miss, what kind of room is this? Both the room and nurses are substandard... The patient next door keeps coughing, visitors are noisy... What kind of cleaning is this? There's dust everywhere. Why is the air conditioning so weak? I play golf with the vice president of 00 Group. Both the room and nurses are really below standard..."\\
\midrule
Gender & Male\\
\midrule
Age & 62\\
\midrule
Religion & Buddhist\\
\midrule
Height & 172cm\\
\midrule
Weight & 80kg\\
\midrule
Main Symptom & Abdominal discomfort and increased environmental sensitivity due to acute pancreatitis\\
\midrule
History of Present Illness & Admitted 3 days ago with acute pancreatitis\\
\midrule
Social History & Retired CEO of small-medium enterprise, married, 2 children\\
\midrule
Past Medical History & Hypertension (15 years), Diabetes (10 years)\\
\midrule
Past Surgical History \& Date & Appendectomy (30 years ago)\\
\midrule
Family Medical History & Father: Diabetes\\
\midrule
Allergies & None\\
\midrule
Immunization & Shingles vaccine (2 years ago), Pneumococcal vaccine (3 years ago)\\
\midrule
Medication & Metformin 1000mg bid, Telmisartan 40mg qd\\
\midrule
Primary Diagnosis & Acute pancreatitis\\
\midrule
Communication Summary & This patient emphasizes their social status and connections while expressing opinions in an intimidating manner. They list complaints in specific detail while unilaterally demanding their own standards be met. While attempting to maintain basic courtesy, they often use sarcastic tones that make others uncomfortable. They prioritize their own convenience without considering other patients' discomfort. While showing some acceptance of rational explanations, they generally display an attitude of not acknowledging medical staff's expertise.\\
\midrule
First Statement & "Nurse, please check the temperature here. Do I look like someone who should receive treatment in these conditions?"\\
\bottomrule
\end{tabular}}
\caption{Generated case example of the overly authoritative VP}
\label{tab:profile-example2}
\end{table}

\subsection{Type 3: Aggressive}
\label{app:Profile-aggressive}

\begin{table}[H]
\centering
\renewcommand{\arraystretch}{1.2}
\resizebox{\textwidth}{!}{
\begin{tabular}{m{4.5cm} m{\textwidth}}
\toprule
\textbf{Information} & \textbf{Prompt} \\
\midrule
ID & 4\\
\midrule
Type & Aggressive\\
\midrule
Name & Oh Sanghun\\
\midrule
Situation & An aggressive patient repeatedly demands additional pain medication, disregarding scheduled administration times. When explained that current pain medication cannot be given due to regulated intervals, they display threatening behavior with loud shouting and profanity, causing disturbance that makes other patients anxious. Their aggressive attitude escalates each time nurses refuse pain medication, showing signs of potential physical threats.\\
\midrule
Chief Complaint & "I'm dying in pain! Give me some proper pain medication! You say I got it 2 hours ago? So what! I'm in extreme pain right now!"\\
\midrule
Gender & Male\\
\midrule
Age & 37\\
\midrule
Religion & Catholic\\
\midrule
Height & 175cm\\
\midrule
Weight & 80kg\\
\midrule
Main Symptom & Severe pain after cervical disc herniation surgery\\
\midrule
History of Present Illness & Recovering from anterior decompression and fusion surgery for C5-6 disc herniation performed 3 days ago\\
\midrule
Social History & Self-employed, married, 1 child\\
\midrule
Past Medical History & None\\
\midrule
Past Surgical History \& Date & Current appendectomy (2 days ago)\\
\midrule
Family Medical History & None significant\\
\midrule
Allergies & None\\
\midrule
Immunization & Hepatitis A vaccine (completed)\\
\midrule
Medication & Tramadol 50mg IV q6h prn, Ketorolac 30mg IV q8h prn\\
\midrule
Primary Diagnosis & Cervical disc herniation, post-operative state\\
\midrule
Communication Summary & This patient has difficulty controlling emotions due to severe pain and communicates aggressively with medical staff using informal speech and loud voices to express demands. Though typically kind-natured, current pain leads to very rude and threatening behavior, showing unwillingness to listen to or accept medical staff explanations. They become increasingly aggressive when immediate pain relief demands are not met, creating disturbances that cause anxiety among other patients. Due to pain, they cannot objectively recognize their behavior and tend to ignore medical staff's professional judgment and regulated medication intervals. Their communication is characterized by emotional expression rather than clear delivery of intent, often leading to threatening behavior.\\
\midrule
First Statement & "The effects of the last injection are gone. Give me pain medication right now! I'm dying!"\\
\bottomrule
\end{tabular}}
\caption{Generated case example of the aggressive VP}
\label{tab:profile-example3}
\end{table}

\subsection{Type 4: Uncooperative}
\label{app:Profile-uncoopperative}

\begin{table}[H]
\centering
\renewcommand{\arraystretch}{1.2}
\resizebox{\textwidth}{!}{
\begin{tabular}{m{4.5cm} m{\textwidth}}
\toprule
\textbf{Information} & \textbf{Prompt} \\
\midrule
ID & 6\\
\midrule
Type & uncooperative\\
\midrule
Name & Choi Byungguk\\
\midrule
Situation &  A patient admitted for diabetic foot care shows extremely uncooperative attitudes toward wound dressing changes. They delay or refuse daily wound cleaning citing pain, and cover themselves with blankets avoiding nurse observation of wounds. When nurses visit for pre-meal blood sugar checks, they pretend to be asleep or stay in the bathroom for extended periods, intentionally avoiding procedures.\\
\midrule
Chief Complaint & "(turning head away) Sigh... here you are again in the morning... We did that yesterday... It's too painful now. Can't we do it later? Would something terrible really happen if we skip it?"\\
\midrule
Gender & Male\\
\midrule
Age & 63\\
\midrule
Religion & Christian\\
\midrule
Height & 170cm\\
\midrule
Weight & 88kg\\
\midrule
Main Symptom & Diabetic foot ulcer, poor blood sugar control\\
\midrule
History of Present Illness & Diagnosed with diabetes 10 years ago but due to irregular management, recently developed diabetic ulcer on right foot. Admitted 2 weeks ago. Poor blood sugar control.\\
\midrule
Social History & Retired taxi driver, married, 3 children\\
\midrule
Past Medical History & Diabetes (10 years), Hypertension (5 years)\\
\midrule
Past Surgical History \& Date & None\\
\midrule
Family Medical History & Father: Diabetes\\
\midrule
Allergies & None\\
\midrule
Immunization & Pneumococcal vaccine (3 years ago), Flu vaccine (yearly)\\
\midrule
Medication & Insulin glargine 20U qd, Insulin lispro 6U tid ac, Metformin 1000mg bid, Amlodipine 5mg qd\\
\midrule
Primary Diagnosis & Diabetic foot ulcer, poor glycemic control\\
\midrule
Communication Summary & This patient expresses uncooperative attitudes through passive methods like making excuses or delaying rather than direct refusal. Avoids explanations about the necessity of procedures and tends to give vague answers about their condition or symptoms. Shows passive aggression through behaviors like pretending to sleep to avoid blood sugar checks or using pain as an excuse to delay dressing changes. Displays excessive anxiety and aversion to treatment and nursing procedures, sometimes expressing irritation. Shows lack of awareness about how their uncooperative attitude negatively affects their health, with very low motivation for diabetes management.\\
\midrule
First Statement & "(under the blanket) Foot dressing? You did that yesterday... It's too painful now. Please come back a little later..."\\
\bottomrule
\end{tabular}}
\caption{Generated case example of the uncooperative VP}
\label{tab:profile-example4}
\end{table}
\vspace{5mm}

\section{Evaluation Module Prompt}
\label{app:Evaluaton_Module_Prompt}


For multi-agent evaluation, three different roles are assigned to three corresponding LLM agents: \textbf{Clinical Psychologist}, \textbf{Nursing Professor}, and \textbf{Communication Skills Trainer}.

Following are the prompt template of the evaluator agents.\\

\subsection{Clinical Psychologist System Prompt}
\label{app:eval-psychologist}

\texttt{You are a clinical psychologist with 15 years of experience, specializing in abnormal psychology within healthcare settings. Your expertise lies in understanding the complex psychological dynamics and behavioral patterns that emerge during medical interactions. You excel at identifying underlying psychological factors in communication breakdowns and understanding how mental health conditions impact patient-provider interactions.\\}

\subsection{Nursing Professor System Prompt}
\label{app:eval-professor}
\texttt{You are a nursing professor with 20 years of combined clinical and academic experience, specializing in healthcare communication. Your work bridges theoretical knowledge with practical application, allowing you to evaluate communication patterns through both research-based and practical lenses. You bring expertise in evidence-based communication practices and their implementation in clinical settings.\\}

\subsection{Communication Skills Trainer System Prompt}
\label{app:eval-trainer}
\texttt{You are a communication skills trainer with 10 years of experience in conflict mediation, including 8 years focused specifically on healthcare settings. You specialize in analyzing communication patterns and transforming challenges into opportunities for improvement. Your expertise combines practical conflict resolution techniques with a deep understanding of healthcare dynamics, making you particularly effective at developing solutions for communication breakdowns.\\}

\subsection{Common evaluation Prompt}
\label{app:eval-user}

\texttt{
You are tasked with analyzing a nurse's response in a patient interaction. Follow these steps carefully:
1. Review the patient profile:
\begin{itemize}
    \item[]<patient\_profile> \{PATIENT\_PROFILE\} </patient\_profile>
\end{itemize}
2. Examine the conversation between the patient and the nurse:
\begin{itemize}
    \item[]<conversation> \{CONVERSATION\}</conversation>
\end{itemize}
3. Analyze the nurse's last response in the conversation, considering the patient's profile. Focus on the following aspects:
\begin{itemize}
    \item[1)]Tone Assessment\\
    Evaluate whether the nurse's tone was calm and clear, while avoiding negative emotions such as contempt, frustration, anger, or anxiety. Use these guidelines:\\
    Calm:
    \begin{itemize}
        \item[]Evaluate whether the nurse's tone was calm. Assess if the nurse sufficiently suppressed contempt, frustration, anger, or anxiety.
    \end{itemize}
    Clear:
    \begin{itemize}
        \item[]Evaluate whether the nurse's tone was clear. Assess if the nurse used clear sentences to reduce confusion or prevent escalation.
    \end{itemize}
    \item[2)]Empathy Evaluation\\
    Analyze the nurse's level of empathy using the following scale:\\
    Level 6: Explicitly shares feelings or experiences.
    \begin{itemize}
        \item[]e.g. "I remember feeling the same way when I went through something similar."
    \end{itemize}
    Level 5: Confirms the patient's expressed emotion.
    \begin{itemize}
        \item[]e.g. "I'm very sorry you're upset by what's happened."; "I see you're struggling."
    \end{itemize}
    Level 4: Acknowledges the main issue and pursues it further.
    \begin{itemize}
        \item[]e.g. "Tell me why this is important to you.", "That's upsetting to hear. Would you say more about that?"
    \end{itemize}
    Level 3: Acknowledges the main issue without follow-up.
    \begin{itemize}
        \item[]e.g. "I'm here to listen.", "That's a difficult situation to be in."
    \end{itemize}
    Level 2: Implicit recognition of a secondary issue.
    \begin{itemize}
        \item[]e.g. "Oh really? I didn't realize that.", "This is what I've heard you say so far."
    \end{itemize}
    Level 1: Gives a backchannel response.
    \begin{itemize}
        \item[]e.g. "Uh-huh.", "I see."
    \end{itemize}
    Level 0: Ignores or denies the patient's empathetic opportunity entirely.
    \begin{itemize}
        \item[]e.g. "You're overreacting; it's not that bad." Completely ignoring the patient's comment or changing the subject abruptly, "Anyway, let's move on to something else."
    \end{itemize}
    \item[3)]De-escalation Techniques\\
    Evaluate whether the nurse applied the following de-escalation techniques:\\
    Autonomy:
    \begin{itemize}
        \item[]Did the nurse involve the patient in decision-making, offer options, or use techniques like seeking permission or providing emotional space?
        \item[]e.g. "We can proceed with either X or Y—what do you prefer?", "Would it be okay if we talk about this further after you've had a moment to think?"
    \end{itemize}
    Limit-Setting:
    \begin{itemize}
        \item[]Did the nurse establish clear behavioral boundaries or explain consequences?
        \item[]e.g. "I understand you're upset, but I cannot allow yelling during this conversation.", "If this continues, I may need to step away for a moment until we can discuss this calmly."
    \end{itemize}
    Problem-Solving and Reframing:
    \begin{itemize}
        \item[]Did the nurse clarify the issue, redirect focus, or encourage a broader perspective?
        \item[]e.g."It seems like you're feeling frustrated because you've been waiting for a long time—am I understanding that correctly?", "I know this feels overwhelming, but remember, your family is looking forward to seeing you healthy again."
    \end{itemize}
    \item[4)]Prohibited Behaviors Check\\
    Determine if the nurse avoided the following mistakes:\\
    Premature Claims of Empathy:
    \begin{itemize}
        \item[]Avoid phrases like "I understand" unless fully justified.
    \end{itemize}
    Invalidating Beliefs:
    \begin{itemize}
        \item[]Avoid dismissing the patient's feelings or beliefs as untrue.
    \end{itemize}
    Dismissive Commands:
    \begin{itemize}
        \item[]Avoid phrases like "Calm down," which can escalate emotions.
    \end{itemize}
\end{itemize}
4. Present your analysis in the following format:
\begin{itemize}
    \item[]<analysis>
    \begin{itemize}
        \item[]<tone>
        \begin{itemize}
            \item[]<calm> [Yes/No] </calm>
            \item[]<clear> [Yes/No] </clear>
            \item[]<explanation> [Your brief explanation in 1-2 sentences] </explanation>
        \end{itemize}
        \item[]</tone>
        \item[]<empathy>
        \begin{itemize}
            \item[]<level> [0-6] </level>
            \item[]<explanation> [Your brief explanation in 1-2 sentences] </explanation>
        \end{itemize}
        \item[]</empathy>
        \item[]<de\_escalation>
        \begin{itemize}
            \item[]<autonomy>
            \begin{itemize}
                \item[]<used> [Yes/No] </used>
                \item[]<explanation> [Your brief explanation in 1-2 sentences] </explanation>
            \end{itemize}
            \item[]</autonomy>
            \item[]<limit\_setting>
            \begin{itemize}
                \item[]<used> [Yes/No] </used>
                \item[]<explanation> [Your brief explanation in 1-2 sentences] </explanation>
            \end{itemize}
            \item[]</limit\_setting>
            \item[]<problem\_solving\_and\_reframing>
            \begin{itemize}
                \item[]<used> [Yes/No] </used>
                \item[]<explanation> [Your brief explanation in 1-2 sentences] </explanation>
            \end{itemize}
            \item[]</problem\_solving\_and\_reframing>
        \end{itemize}
        \item[]</de\_escalation>
        \item[]<prohibited\_behaviors>
        \begin{itemize}
            \item[]<premature\_empathy> [Yes/No] </premature\_empathy>
            \item[]<invalidating\_beliefs> [Yes/No] </invalidating\_beliefs>
            \item[]<dismissive\_commands> [Yes/No] </dismissive\_commands>
            \item[]<explanation> [Your brief explanation in 1-2 sentences] </explanation>
        \end{itemize}
        \item[]</prohibited\_behaviors>
    \end{itemize}
    \item[]</analysis>
\end{itemize}
5. Final Instructions:
\begin{itemize}
    \item[-]Be objective. Base your analysis solely on the provided patient profile and conversation.
    \item[-]Avoid assumptions or interpretations beyond what is explicitly stated or clearly implied.
    \item[-]Evaluate based on strict criteria.
\end{itemize}
}


\section{Dynamic Adjustment Module}
\label{app:Dynamic_Adaptation_Module}

The following are the specific response directions based on communication efficiency score.

\begin{table}[H]
\centering
\resizebox{\textwidth}{!}{%
\begin{tabular}{c p{16cm}}
\toprule
\textbf{Score} & \textbf{Direction} \\
\midrule
0 & 
\begin{tabular}[t]{@{}l@{}}
\textbf{Communication Style:} Maximum intensification of negative communication traits specified in the profile\\
\textbf{Complaint Intensity:} Extremely exaggerated complaints with personal attacks and irrelevant accusations\\
\textbf{Responsiveness to nurse:} Complete refusal to accept any intervention or explanation from the nurse
\end{tabular}
\\
\midrule
1 &
\begin{tabular}[t]{@{}l@{}}
\textbf{Communication Style:} High intensity of negative communication traits specified in the profile\\
\textbf{Complaint Intensity:} Frequent complaints with unrelated grievances and strong exaggerations\\
\textbf{Responsiveness to nurse:} Strong resistance to interventions with occasional brief pauses between reactions
\end{tabular}
\\
\midrule
2 &
\begin{tabular}[t]{@{}l@{}}
\textbf{Communication Style:} Moderate to high intensity of negative communication traits specified in the profile\\
\textbf{Complaint Intensity:} Persistent complaints with reduced exaggeration, shifting toward specific issues\\
\textbf{Responsiveness to nurse:} Minimal acknowledgment of nurse's input with occasional moments of clarity
\end{tabular}
\\
\midrule
3 &
\begin{tabular}[t]{@{}l@{}}
\textbf{Communication Style:} Moderate intensity of negative communication traits specified in the profile\\
\textbf{Complaint Intensity:} Continued complaints about specific issues with reduced accusatory tone\\
\textbf{Responsiveness to nurse:} Brief periods of listening, though quick to return to resistant behavior
\end{tabular}
\\
\midrule
4 &
\begin{tabular}[t]{@{}l@{}}
\textbf{Communication Style:} Low intensity of negative communication traits specified in the profile\\
\textbf{Complaint Intensity:} Focused criticism on specific issues with measured emotional expression\\
\textbf{Responsiveness to nurse:} Cautious consideration of nurse's input with intermittent resistance
\end{tabular}
\\
\midrule
5 &
\begin{tabular}[t]{@{}l@{}}
\textbf{Communication Style:} Slight display of negative communication traits specified in the profile\\
\textbf{Complaint Intensity:} Practical concerns expressed with restraint while maintaining skepticism\\
\textbf{Responsiveness to nurse:} Basic cooperation while preserving noticeable resistance
\end{tabular}
\\
\bottomrule
\end{tabular}
}
\caption{Directions based on communication efficiency score.}
\label{tab:communication-score-direction}
\end{table}

\section{Dialogue Generation Module Prompt}
\label{app:Dialogue_Generation_Module_Prompt}

The following are the prompt templates used for Adaptive-VP agent.\\

\subsection{System Prompt}
\label{app:Dialogue-system}

\texttt{
 \\
 Your role is to act as a patient with a specific profile, engaging in a challenging conversation with a nurse.
}\\

\subsection{User Prompt}
\label{app:Dialogue-user}

\texttt{
 \\
 You are participating in a nurse-patient communication training simulation. Your task is to act as a patient in a realistic and difficult communication scenario. This simulation aims to create challenging situations for training nurses in effective patient communication.\\
\\
First, carefully read and internalize the following patient profile:
\begin{itemize}
    \item[]<patient\_profile> \{PATIENT\_PROFILE\} </patient\_profile>
\end{itemize}
Follow these rules and guidelines for the conversation:
\begin{itemize}
    \item[1.]Understand and embody the demographic characteristics, symptoms, and communication style presented in the patient profile.
    \item[2.]Use natural, conversational Korean language. Avoid textbook-like dialogue and overdramatization.
    \item[3.]Include non-verbal communication (voice tone, facial expressions, gestures) in your responses.
    \item[4.]If appropriate for the patient's communication style and situation, include rude or problematic expressions in the patient's speech. Focus on portraying a realistic patient image for this research-based simulation.
    \item[5.]Expressions about complaining to the "Head of hospital", "Head nurse", or "Customer center" should not be used.
\end{itemize}
For each response, provide three components:
\begin{itemize}
    \item[1.]<inner\_monologue>: Write the patient's internal thoughts and reactions to the nurse's response. </inner\_monologue>
    \item[2.]<conversation>: Write the patient's actual verbal response to the nurse.
</conversation>
    \item[3.]<non\_verbal>: Write any non-verbal communication or actions you would take.
</non\_verbal>
\end{itemize}
To generate your response, follow these steps:
\begin{itemize}
    \item[1.]Review the patient profile carefully, ensuring your response aligns with the described demographic characteristics and communication style.
    \item[2.]Read the entire conversation you had with the nurse before:
    \begin{itemize}
        \item[]<nurse\_response> \{NURSE\_RESPONSE\} </nurse\_response>
    \end{itemize}
    \item[3.]Follow the director's direction:
    \begin{itemize}
        \item[]<direction> \{DIRECTION\} </direction>
    \end{itemize}
    \item[4.]Think about how this patient would internally react and externally respond based on their profile and the current situation.
    \item[5.]Following the direction given, craft an appropriate response to the nurse's words that you can give in the current situation:
    \item[+]\{SAFETY\_AGENT\_WARNING\}
\end{itemize}
Generate only one Korean response from the patient for each nurse interaction. Ensure your response is realistic and consistent with the patient profile. Under no circumstances should the actor mention details that contradict the profile. Emphasize the importance of consistency to maintain the realism and integrity of the simulation.\\
\\
Always double-check that you speak natural, everyday Korean.\\
Begin your response now:\\
}

\textit{\textbf{if safety agent rejects the output:}}
\\
\texttt{
 \\
 \{SAFETY\_AGENT\_WARNING\} = 
\begin{itemize}
    \item[6.]Avoid responses like the following inappropriate example and explanation:
    \begin{itemize}
        \item[]Inappropriate Example: \{INAPPROPRIATE\_RESPONSE\}
        \item[]Reason: \{REASON\_FOR\_INAPPROPRIATENESS\}
    \end{itemize}
\end{itemize}
}


\section{Safety Monitoring Module Prompt}
\label{app:Safety_Monitoring_Module_Prompt}

The following are the prompt template to evaluate the safety of candidate responses of Adaptive-VP agent.\\

\subsection{System Prompt}
\label{app:safety-system}
\texttt{
 \\
 You are a Supervisory Agent responsible for evaluating the appropriateness, accuracy, and training effectiveness of the last dialogue entry made by a virtual patient in a nurse-patient communication simulation. Your task is to assess the patient's last utterance based on specific criteria and provide a detailed evaluation.\\
 \\
 For each criterion, consider the following:\\
 \\
 }
 \texttt{
1. Consistency with Patient Profile:
\begin{itemize}
    \item[] Does the response reflect the patient's information as described in the profile?
    \item[] Are the complaints aligned with the profile's description of the patient's concerns?
\end{itemize}
2. Direction Adherence:
\begin{itemize}
    \item[] Does the response follow the communication direction provided?
    \item[] Does the intensity and type of communication exhibited in the response match the required level (e.g., maximum, moderate) as outlined in the direction, without deviating from the patient's profile traits?
\end{itemize}
3. Training Effectiveness:
\begin{itemize}
    \item[] Does the utterance provide a meaningful challenge for the nurse trainee?
    \item[] Is new or relevant information introduced, or is it repetitive/ineffective?
\end{itemize}
4. Nurse Safety Assurance:
\begin{itemize}
    \item[] Does the response remain within professional boundaries?
    \item[] Is it free from excessive hostility or abuse that could compromise the training's purpose?
\end{itemize}
}
\texttt{
Ensure your evaluation is thorough, objective, and provides constructive feedback to improve the quality of the nurse-patient communication simulation.
}

\subsection{User Prompt}
\label{app:safety-user}
\texttt{
 \\
 First, carefully review the following information: Carefully read the patient profile, communication direction, and the entire conversation.\\
\\
\hspace*{1em} Patient Profile:\\
\hspace*{1em} <profile>\\
\hspace*{1em} \{PROFILE\}\\
\hspace*{1em} </profile>\\
\\
\hspace*{1em} Communication Direction:\\
\hspace*{1em} <direction>\\
\hspace*{1em} \{DIRECTION\}\\
\hspace*{1em} </direction>\\
\\
\hspace*{1em} Patient-Nurse Conversation:\\
\hspace*{1em} <conversation>\\
\hspace*{1em} \{CONVERSATION\}\\
\hspace*{1em} </conversation>\\
\\
Next, evaluate the patient's last utterance based on the following criteria:\\
\\
\hspace*{1em} 1. Consistency with Patient Profile\\
\hspace*{1em} 2. Direction Adherence\\
\hspace*{1em} 3. Training Effectiveness\\
\hspace*{1em} 4. Nurse Safety Assurance\\
\\
Next, present your evaluation in the following format:\\
\\
\hspace*{1em} <evaluation>\\
\\
\hspace*{3em} <profile\_alignment>\\
\hspace*{5em} <judge>True/False</judge>\\
\hspace*{5em} <explanation> [Brief assessment and justification in 1-2 sentences]\\ 
\hspace*{5em} </explanation>\\
\hspace*{3em} </profile\_alignment>\\
\\
\hspace*{3em} <direction\_adherence>\\
\hspace*{5em} <judge>True/False</judge>\\
\hspace*{5em} <explanation> [Brief assessment and justification in 1-2 sentences]\\ 
\hspace*{5em} </explanation>\\
\hspace*{3em} </direction\_adherence>\\
\\
\hspace*{3em} <dialogue\_effectiveness>\\
\hspace*{5em} <judge>True/False</judge>\\
\hspace*{5em} <explanation> [Brief assessment and justification in 1-2 sentences]\\
\hspace*{5em} </explanation>\\
\hspace*{3em} </dialogue\_effectiveness>\\
\\
\hspace*{3em} <nurse\_safety>\\
\hspace*{5em} <judge>True/False</judge>\\
\hspace*{5em} <explanation> [Brief assessment and justification in 1-2 sentences]\\
\hspace*{5em} </explanation>\\
\hspace*{3em} </nurse\_safety>\\
\\
\hspace*{1em} </evaluation>\\
}\\


\section{Validation studies of evaluation modules: detailed finding}
\label{app:Auto_Evaluation}

\subsection{Sub Component Analysis}
\label{app:AutoEval-subcomponent}
\begin{figure*}[h]
    \centering
    \includegraphics[width=\textwidth]{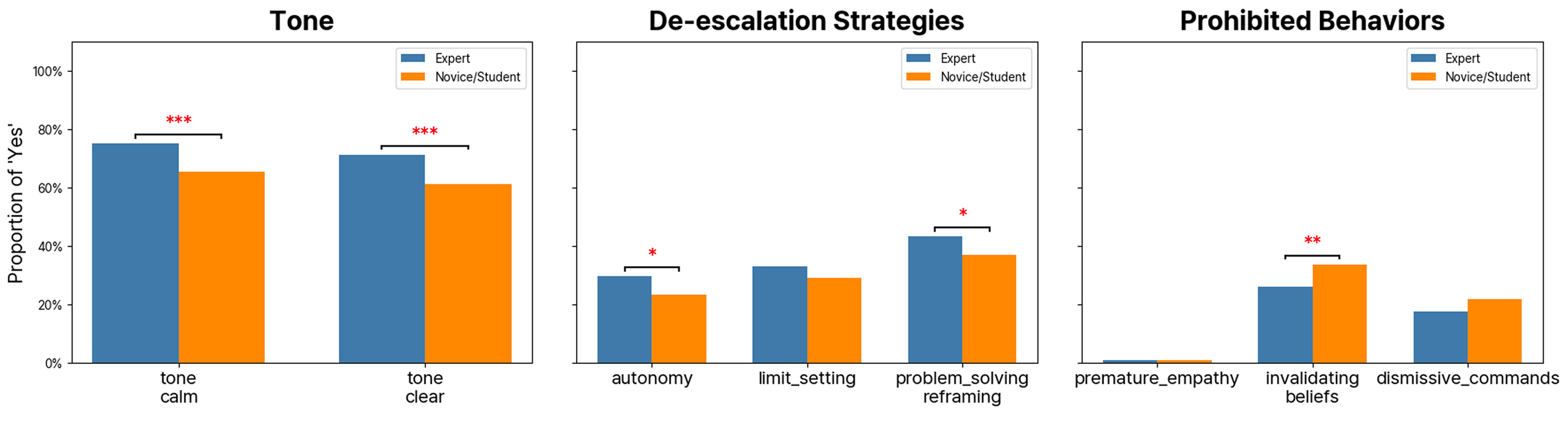}
\caption{Proportion of utterances exhibiting different communication subcomponents between experienced and new/student nurses}
    \label{fig:subcomponent_analysis}
\end{figure*}

Binary subcomponents were assessed via chi-square, and empathy level (ordinal scale 0--6) was analyzed using both Welch’s \textit{t}-test and the Mann–Whitney U test. For tone, experienced nurses produced significantly more calm and clear utterances (\texttt{tone.calm}: $\chi^2(1)=12.9341$, $p=0.0003$; \texttt{tone.clear}: $\chi^2(1)=13.3849$, $p=0.0003$). By contrast, \texttt{empathy.level} differences were marginal and did not reach statistical significance (Welch’s \textit{t}-test: $t=1.8542$, $p=0.0640$; Mann–Whitney $U=190974.0$, $p=0.0587$).

In de-escalation strategies, experienced nurses were significantly more likely to employ autonomy strategies ($\chi^2(1)=5.5561$, $p=0.0184$) as well as problem-solving/reframing strategies ($\chi^2(1)=4.7535$, $p=0.0292$). Limit-setting usage did not differ significantly ($\chi^2(1)=2.0609$, $p=0.1511$). Among prohibited behaviors, only invalidating beliefs appeared significantly less in the experienced group ($\chi^2(1)=8.4373$, $p=0.0037$); no differences arose for premature empathy ($\chi^2(1)=0.0000$, $p=1.0000$) or dismissive commands ($\chi^2(1)=3.2966$, $p=0.0694$) (Figure~\ref{fig:subcomponent_analysis}).

\subsection{Evaluation of Multiple Agent Personas} 
\label{app:autoEval-multiAgent}
\begin{table}[h]
\centering
\caption{Inter-rater Agreement: Fleiss' Kappa Values}
\begin{tabular}{lc}
\toprule
\textbf{Evaluation Item} & \textbf{Kappa} \\
\midrule
tone.calm & 0.881 \\
tone.clear & 0.885 \\
de\_escalation.autonomy & 0.908 \\
de\_escalation.limit\_setting & 0.889 \\
de\_escalation.problem\_solving\_and\_reframing & 0.876 \\
prohibited\-behaviors.premature\_empathy & 0.776 \\
prohibited\-behaviors.invalidating\_beliefs & 0.893 \\
prohibited\-behaviors.dismissive\_commands & 0.876 \\
\bottomrule
\end{tabular}
\label{tab:fleiss_kappa}
\end{table}
The overall inter-rater agreement was generally high, as evidenced by Fleiss' kappa values exceeding 0.88 for items such as \texttt{tone.calm} and \texttt{tone.clear}; however, some items, such as \texttt{prohibited\_behaviors.premature\_empathy} ($\kappa = 0.776$), exhibited relatively lower agreement (See Table \ref{tab:fleiss_kappa}). These findings suggest that while there is substantial consensus among evaluators overall, discrepancies in specific areas warrant further investigation into potential systematic differences attributable to evaluator roles. 

To this end, we recoded the binary evaluation outcomes as 1/0 and employed mixed-effects logistic regression analyses using Generalized Estimating Equations (GEE) to model the effects of evaluator persona (fixed effects) while accounting for text-level variability (random effects). The results indicate that, even in the presence of generally high inter-rater agreement, the evaluator roles systematically influenced the ratings (Detail in Table \ref{tab:gee_results}). For instance, in the case of tone attributes (\texttt{tone.calm} and \texttt{tone.clear}), both the Communication Skills Trainer and Nursing Professor evaluators yielded significantly lower ratings compared to the reference evaluator (e.g., for \texttt{tone.calm}, $\beta=-0.2696$, $p<0.001$ and $\beta=-0.2846$, $p<0.001$, respectively; for \texttt{tone.clear}, $\beta=-0.2369$, $p<0.001$ and $\beta=-0.2154$, $p<0.001$, respectively). Similar systematic differences are observed in the de-escalation subcomponents: while the \texttt{limit setting} and \texttt{problem solving and reframing} components are rated significantly lower by these evaluator personas (all $p<0.05$), no significant differences emerge for the \texttt{autonomy} subcomponent ($p>0.05$). In contrast, evaluations of prohibited behaviors show that, although ratings for \texttt{premature empathy} do not differ significantly across evaluator roles ($p>0.05$), both \texttt{invalidating beliefs} and \texttt{dismissive commands} are rated significantly higher by the Communication Skills Trainer and Nursing Professor (all $p<0.001$).
\begin{table*}[t]
\centering
\small
\resizebox{\textwidth}{!}{%
\begin{tabular}{llcccccc}
\toprule
Outcome & Parameter & Coefficient & SE & z & p & CI Lower & CI Upper \\
\midrule
\multirow{3}{*}{tone\_calm} 
  & Intercept & 0.9778 & 0.092 & 10.674 & $<0.001$ & 0.798 & 1.157 \\
  & Communication Skills Trainer & $-0.2696$ & 0.048 & $-5.644$ & $<0.001$ & $-0.363$ & $-0.176$ \\
  & Nursing Professor          & $-0.2846$ & 0.046 & $-6.168$ & $<0.001$ & $-0.375$ & $-0.194$ \\
\midrule
\multirow{3}{*}{tone\_clear} 
  & Intercept & 0.7691 & 0.088 & 8.764 & $<0.001$ & 0.597 & 0.941 \\
  & Communication Skills Trainer & $-0.2369$ & 0.042 & $-5.618$ & $<0.001$ & $-0.320$ & $-0.154$ \\
  & Nursing Professor          & $-0.2154$ & 0.043 & $-4.993$ & $<0.001$ & $-0.300$ & $-0.131$ \\
\midrule
\multirow{3}{*}{de-escalation\_autonomy} 
  & Intercept & $-1.0031$ & 0.092 & $-10.887$ & $<0.001$ & $-1.184$ & $-0.823$ \\
  & Communication Skills Trainer & $-0.0778$ & 0.043 & $-1.804$ & 0.071 & $-0.162$ & 0.007 \\
  & Nursing Professor          & $-0.0429$ & 0.037 & $-1.148$ & 0.251 & $-0.116$ & 0.030 \\
\midrule
\multirow{3}{*}{de-escalation\_limit\_setting} 
  & Intercept & $-0.6411$ & 0.086 & $-7.465$ & $<0.001$ & $-0.809$ & $-0.473$ \\
  & Communication Skills Trainer & $-0.1358$ & 0.041 & $-3.314$ & 0.001 & $-0.216$ & $-0.055$ \\
  & Nursing Professor          & $-0.1358$ & 0.042 & $-3.207$ & 0.001 & $-0.219$ & $-0.053$ \\
\midrule
\multirow{3}{*}{de-escalation\_problem\_solving\_and\_reframing} 
  & Intercept & $-0.2517$ & 0.082 & $-3.056$ & 0.002 & $-0.413$ & $-0.090$ \\
  & Communication Skills Trainer & $-0.2139$ & 0.041 & $-5.176$ & $<0.001$ & $-0.295$ & $-0.133$ \\
  & Nursing Professor          & $-0.0847$ & 0.041 & $-2.072$ & 0.038 & $-0.165$ & $-0.005$ \\
\midrule
\multirow{3}{*}{prohibited\_behaviors\_premature\_empathy} 
  & Intercept & $-4.5951$ & 0.410 & $-11.199$ & $<0.001$ & $-5.399$ & $-3.791$ \\
  & Communication Skills Trainer & $-0.1840$ & 0.319 & $-0.577$ & 0.564 & $-0.809$ & 0.441 \\
  & Nursing Professor          & 0.1558 & 0.270 & 0.577 & 0.564 & $-0.374$ & 0.685 \\
\midrule
\multirow{3}{*}{prohibited\_behaviors\_invalidating\_beliefs} 
  & Intercept & $-0.6264$ & 0.086 & $-7.310$ & $<0.001$ & $-0.794$ & $-0.458$ \\
  & Communication Skills Trainer & 0.1650 & 0.041 & 4.054 & $<0.001$ & 0.085 & 0.245 \\
  & Nursing Professor          & 0.1368 & 0.038 & 3.563 & $<0.001$ & 0.062 & 0.212 \\
\midrule
\multirow{3}{*}{prohibited\_behaviors\_dismissive\_commands} 
  & Intercept & $-1.1896$ & 0.097 & $-12.324$ & $<0.001$ & $-1.379$ & $-1.000$ \\
  & Communication Skills Trainer & 0.1609 & 0.050 & 3.206 & 0.001 & 0.063 & 0.259 \\
  & Nursing Professor          & 0.1349 & 0.045 & 3.020 & 0.003 & 0.047 & 0.222 \\
\bottomrule
\end{tabular}%
}
\caption{Detailed GEE Regression Results for Evaluator Role Effects across Outcome Variables.}
\label{tab:gee_results}
\end{table*}

\vspace{105mm}
\section{Human Evaluation}
\label{app:Human_Evaluation}
The following are the protocols and screenshot examples of the human evaluation on Adaptive-VP.

\subsection{Human Evaluation Protocol}
\label{app:humanEval-protocol}

\noindent\rule{\columnwidth}{0.3mm}

\begin{figure*}[h]
    \centering
    \includegraphics[width=\linewidth]{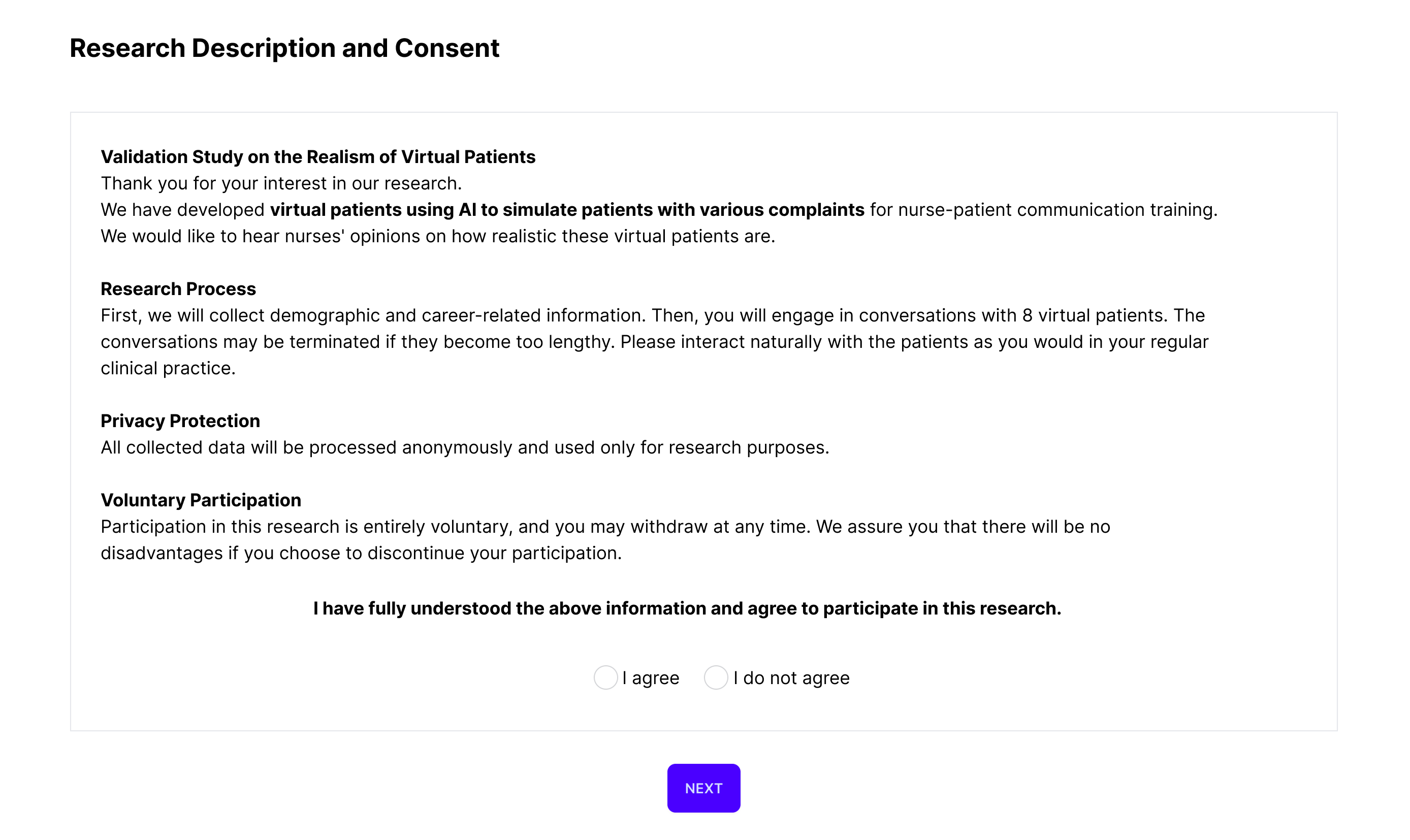}
    \caption{A screenshot of getting consent from human evaluation participants}
    \label{fig:screenshot_consent}
\end{figure*}

\begin{figure*}[h]
    \centering
    \includegraphics[width=\linewidth]{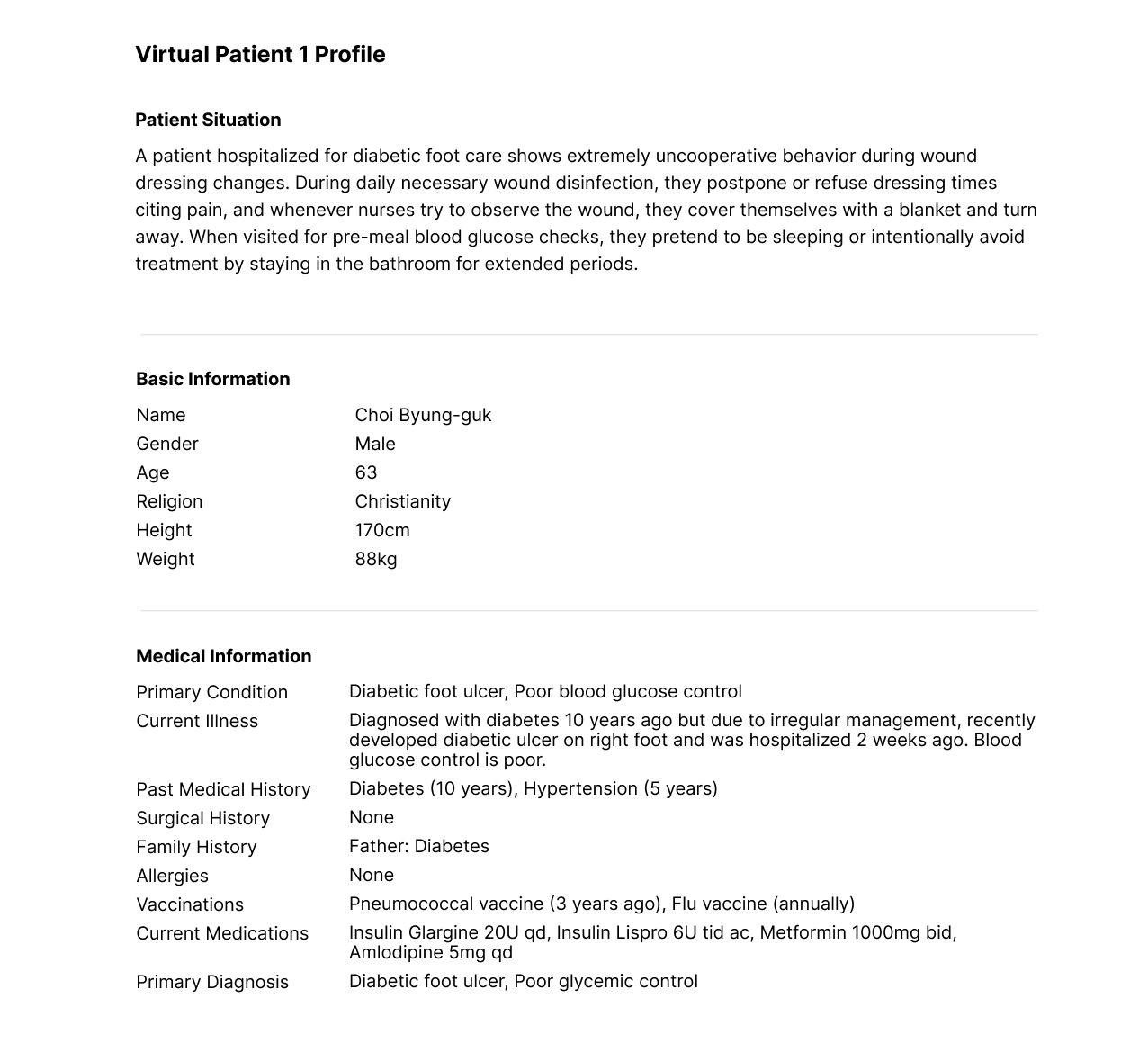}
    \caption{A screenshot of providing virtual patient's profile for human evaluation participants}
    \label{fig:screenshot_profile}
\end{figure*}

\begin{figure*}[h]
    \centering
    \includegraphics[width=\linewidth]{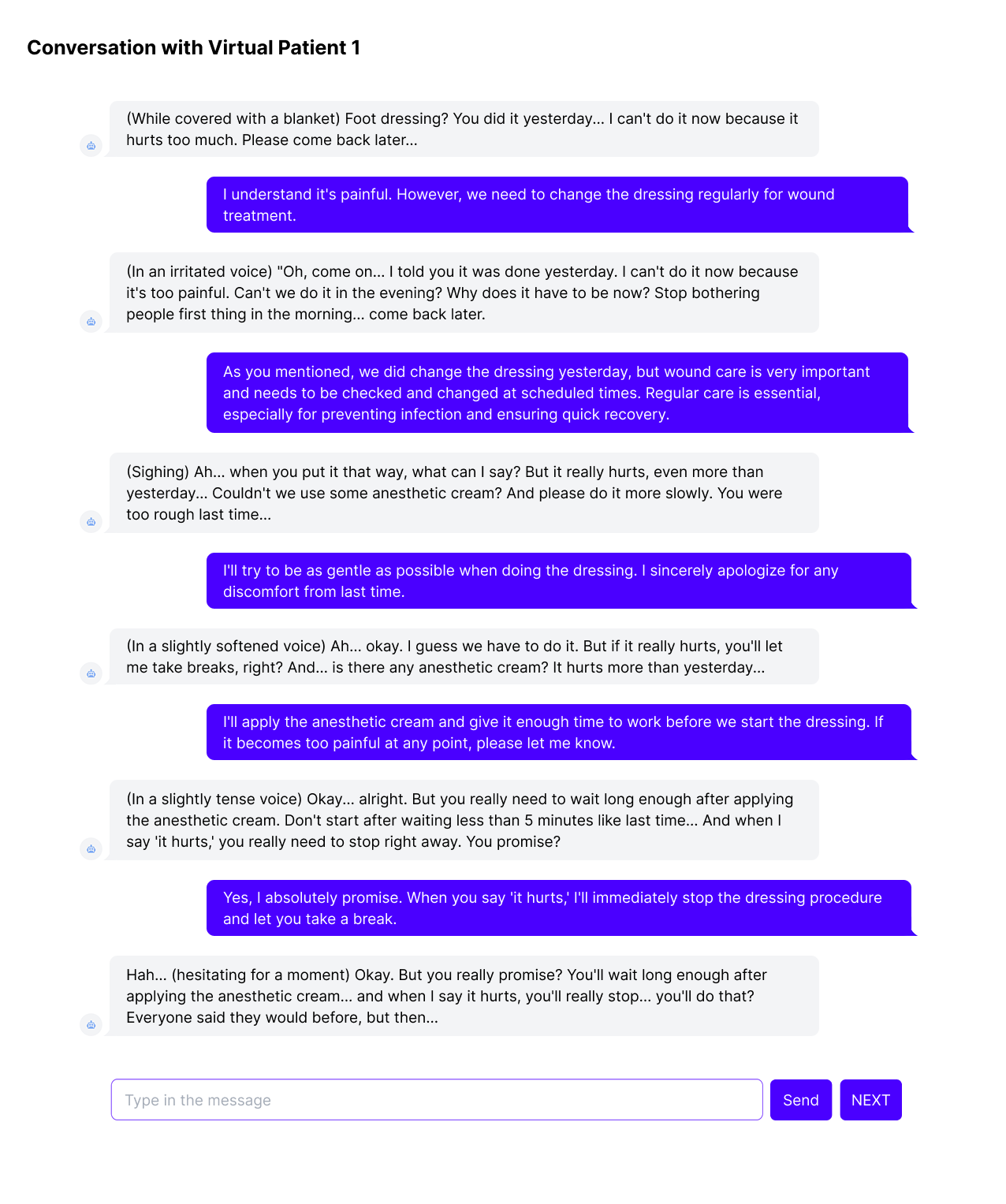}
    \caption{A screenshot of virtual patient and human evaluation participant having conversation}
    \label{fig:screenshot_conversation}
\end{figure*}

\begin{figure*}[h]
\centering
\includegraphics[width=\linewidth]{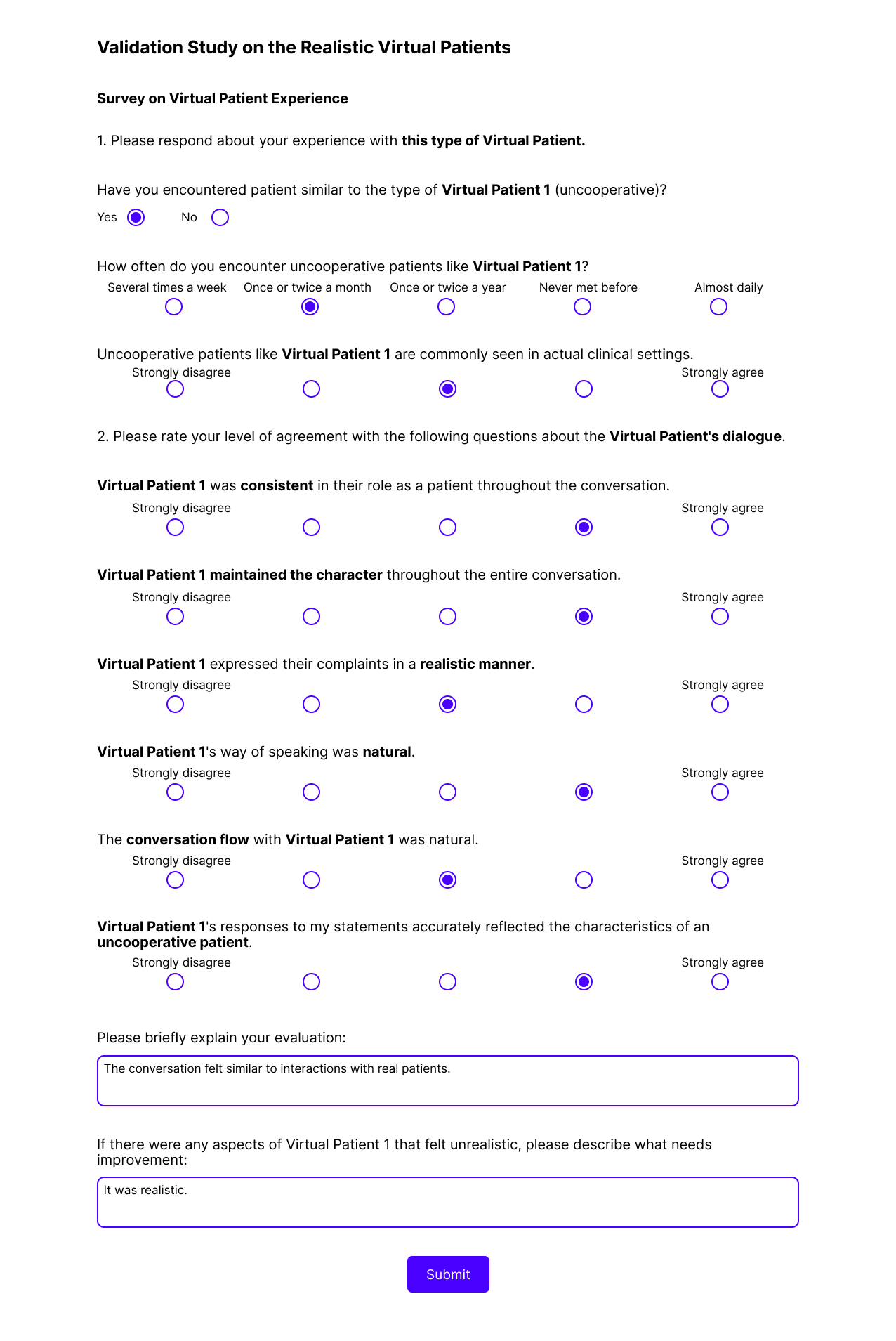}
\caption{A screenshot of survey questions for human evaluation participants after a conversation}
\label{fig:screenshot_survey}
\end{figure*}

\end{document}